\pdfoutput=1
\documentclass[11pt]{article}

\usepackage[]{acl}

\usepackage{times}
\usepackage{latexsym}
\usepackage{graphicx}
\usepackage{ amssymb }
\usepackage{caption}
\usepackage{multirow}
\usepackage{subcaption}
\usepackage[T1]{fontenc}
\usepackage{color}
\usepackage{mathtools}

\usepackage[utf8]{inputenc}

\usepackage{microtype}
\usepackage{url}
\usepackage{amsmath}
\usepackage{tabularx}
\usepackage{longtable}
\usepackage{arydshln}
\usepackage{xspace}
\usepackage[inline]{enumitem}

\title{Knowledge-Centric Templatic Views of Documents}

\author{Isabel Cachola\thanks{\xspace This work was conducted by the first author during an internship at Microsoft Research.} $^\text{1}$  Silviu Cucerzan$^\text{2}$ Allen Herring$^\text{2}$ \\
\textbf{Vuksan Mijovic$^\text{2}$  Erik Oveson$^\text{2}$ Sujay Kumar Jauhar$^\text{2}$} \\
$^1$Johns Hopkins University  $^2$Microsoft \\
\texttt{icachola@cs.jhu.edu},\\
\texttt{\{silviu,allenh,vmijovic,erikov,sjauhar\}@microsoft.com}}

\begin{document}
\newcommand\isabel[1]{{\color{purple}\{#1\}$_{Isabel}$}}

\newcommand\todoit[1]{{\color{red}\{TODO: \textit{#1}\}}}
\newcommand\todo{{\color{red}{TODO}}\xspace}
\newcommand\todocite{{\color{red}{CITE}}\xspace}
\newcommand{\red}[1]{\textcolor{red}{#1}} 
\newcommand{\blue}[1]{\textcolor{blue}{#1}} 
\maketitle

\begin{abstract}
Authors seeking to communicate with broader audiences often share their ideas in various document formats, such as slide decks, newsletters, reports, and posters.
Prior work on document generation has generally tackled the creation of each separate format to be a different task, leading to fragmented learning processes, redundancy in models and methods, and disjointed evaluation. We consider each of these documents as \textit{templatic views} of the same underlying knowledge/content, and we aim to unify the generation and evaluation of these templatic views. We begin by showing that current LLMs are capable of generating various document formats with little to no supervision. Further, a simple augmentation involving a structured intermediate representation can improve performance, especially for smaller models. We then introduce a novel unified evaluation framework that can be adapted to measuring the quality of document generators for heterogeneous downstream applications. This evaluation is adaptable to a range of user defined criteria and application scenarios, obviating the need for task specific evaluation metrics. Finally, we conduct a human evaluation, which shows that people prefer 82\% of the documents generated with our method, while correlating more highly with our unified evaluation framework than prior metrics in the literature.

\end{abstract}

% \sjauhar{Minor request: can we have the different sections in different files? Easier to keep track of things and go to relevant parts of the doc.}
\begin{figure*}[!t]
    \centering
    \includegraphics[width=.8\textwidth]{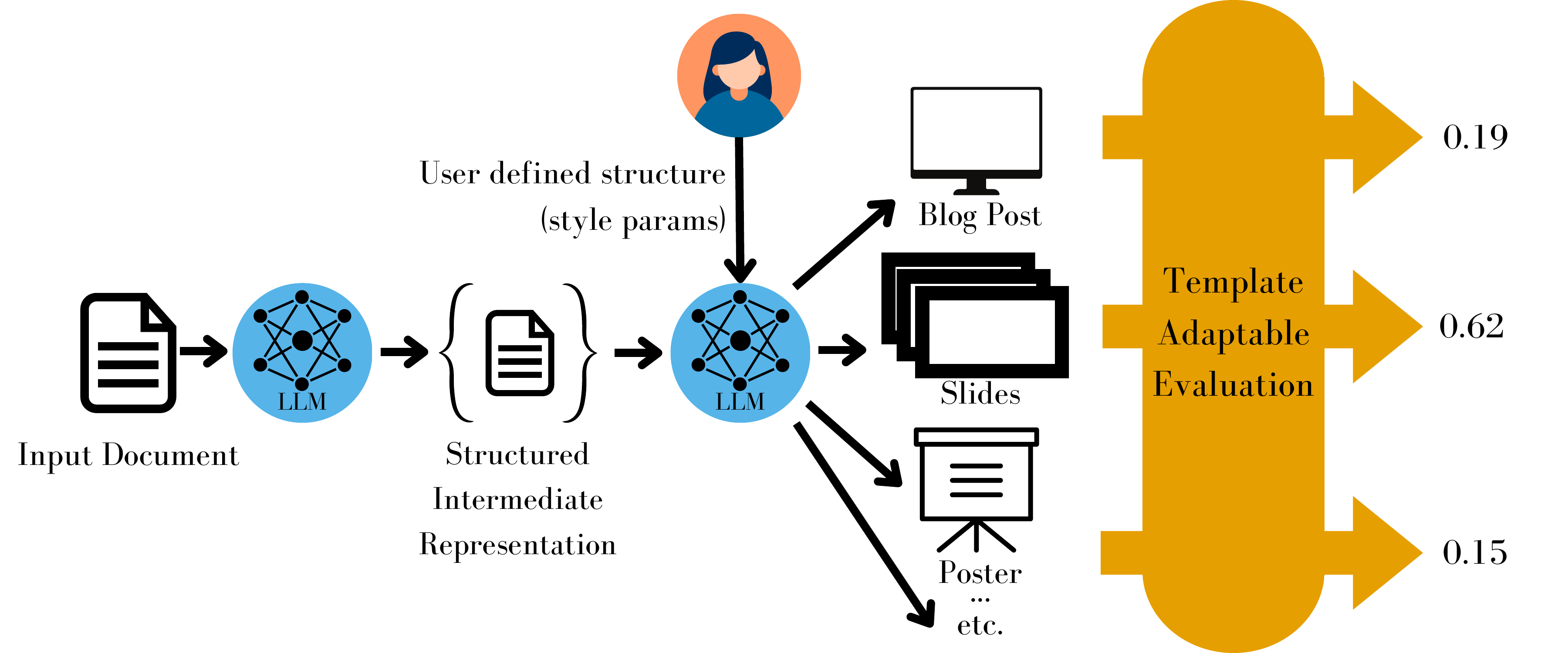}
    \caption{\small Visualization of our method to unify the generation and evaluation of templatic views of documents. Given an input document, we prompt the LLM to generate an intermediate representation. We can use the representation to prompt the model to generate a templatic view of the input document. We then evaluate the generations using our unified evaluation framework. The LLM represented in the figure is the same model.}
    \label{fig:method}
    \vspace{-2mm}
\end{figure*}
\section{Introduction}
Sharing information is vital for communication and discourse across domains, as it allows for knowledge to be disseminated to a wider audience. This is often done by users through documents in multiple formats that nevertheless share some underlying knowledge. A product manager may need to create a requirements spec, a product pitch deck, and an announcement newsletter for the same project. Likewise, a person on the job market may create a resume, a cover letter, and a personal website. We consider these documents to be \textit{templatic views} of the same underlying knowledge.

This is equally true for the scientific domain, in which researchers create documents in multiple formats to effectively communicate and showcase their work, -- such as through academic papers, conference talks, social media posts, poster presentations, and non-technical blog posts. Sharing knowledge in multiple formats broadens the audience and can help bridge the information gap between domain experts, researchers in adjacent fields, and even the general public, leading to greater understanding, collaborations and accelerated progress~\cite{Bornmann2014GrowthRO}.
%With the significant increase in the pace of publication in recent years~\cite{Bornmann2014GrowthRO}, it has become essential for scientists to engage with the community and share their work in more compact, efficient, or easier to digest formats. 
% Not only it can help increase public understanding of scientific topics but also reduce the burden of staying up-to-date in the scientific community. The same holds true for other domains. 

% \sjauhar{I'd start by motivating this even more broadly; nothing about this approach is theoretically limited to the scientific literature and modalities that researchers care about. You can then narrow down to say we focus on a limited domain for reasons...}. 
% Each templatic view reflects different styles and formats. 
% Typically, slide decks employ concise text in bullet point form instead of full sentences, while blog posts are organized into relatively paragraphs with full sentences, which address the most important findings in the work. 
Past work on document generation has focused on developing generation and evaluation methods specific to a single document type~\cite{Fu2021DOC2PPTAP,Qiang2016LearningTG,chandrasekaran-etal-2020-overview-insights}. 
Narrow, custom methods tailored to individual document types are, nevertheless, time consuming to engineer and manage over the long term. For example, in an enterprise setting, it’s common to have dozens of occupation- and task-specific documents, each with their own template.% It is impractical to manage task specific methods for each of these templates, which often evolve over time.
Additionally, specific trained methods require data that may be expensive to acquire, or even be unavailable entirely. Meanwhile, LLMs have recently shown great success in long document generation~\cite{Radford2019LanguageMA, Brown2020LanguageMA}, indicating that this fragmentation of methods may no longer be necessary. Thus, our goal is to unify methods for both generating and evaluating templatic views of documents, allowing system designers and engineers to manage and adapt to a range of document types and domains easily and efficiently. 
% \sjauhar{Why is unification even important? Why should the research community care? Briefly explain; e.g. efficiency in productionization, porting to new domains, etc. More broadly how is \emph{research} in this area going to be beneficial to NLP?}. 
%Our goal is for rur system must be unified, adaptable, and require minimal supervision.  As each document template has different structural and stylistic norms, the method must be able to follow the norms associated with each template. 

We begin by showing that LLMs are capable of diverse, structured document generation, requiring very little instructional guidance to do so effectively. Additionally, a few minor augmentations to the prompt -- such as a structured, intermediate representation, and simple stylistic descriptions -- can further improve downstream performance, especially for smaller, less resource intensive models. These findings have important implications on the deployment and scaling of unified, real-world AI-assisted document authoring systems.

In similar vein, we then introduce Template Adaptable Evaluation (TAE), departing from prior work's task specific evaluation methods~\cite{Zhang2019BERTScoreET, Qiang2016LearningTG,Wang2015iPosterIP}. TAE is a unified precision-recall style framework for automatic evaluation that is highly customizable, allowing users to easily integrate existing text-based metrics from the literature into its formulation and tailor it to their specific use case.% This adaptability includes styles of documents for which there is no publicly available data, such as enterprise-internal product documents. 
Additionally, this framework allows developers to compare performance across document types, without needing to develop an evaluation metric for each individual template.

%Similar to generation methodology, prior work has developed specific evaluation methods for each task~\cite{Zhang2019BERTScoreET, Qiang2016LearningTG,Wang2015iPosterIP}. In this paper, we introduce Template Adaptable Evaluation (TAE) -- a unified precision-recall style framework for automatic evaluation that is highly customizable, allowing users to easily integrate existing text-based metrics from the literature into its formulation and tailor it to their specific use case. This adaptability includes styles of documents for which there is no publicly available data, such as enterprise-internal product documents. Additionally, this framework allows developers to compare performance across document types, without needing to develop an evaluation metric for each individual template. 

% \sjauhar{There needs to be a bit more description of the evaluation setup. What are the datasets and problems?}

We evaluate our unified approach for templatic view generation and evaluation on 3 types of documents: slides, posters, and blog posts~\cite{Fu2021DOC2PPTAP,Qiang2016LearningTG,chandrasekaran-etal-2020-overview-insights}. %We use the DOC2PPT~\cite{Fu2021DOC2PPTAP}, Paper-Poster~\cite{Qiang2016LearningTG}, and LongSumm~\cite{chandrasekaran-etal-2020-overview-insights} datasets, respectively, where the inputs are scientific articles and the outputs are the documents of the corresponding types.
Our experiments demonstrate that using a structured intermediate representation leads to improvements in performance across tasks, with greater gains for smaller language language models.
%demonstrating that LLMs are capable of generating templatic views with structure-augmented prompts, despite having no supervision.
In our human evaluation to validate both our unified document generation method and evaluation metric, we show that annotators prefer the output yielded by the structure-aware generation process 82\% of the time and that our evaluation metric correlates more highly with human preference than other popular metrics. We release our code\footnote{\url{https://github.com/microsoft/knowledge-centric-templatic-views}} to support future research.

%\sjauhar{This feels quite future work-ish; maybe consider moving to Conclusion? I wrote the alternate version above.} While our method focuses on a one-to-many setup (generate the content of multiple templatic views from an input document), our work is a step towards a larger goal of generalized document generation, or more specifically, the many-to-many setup, in which we must synthesize the information contained in multiple documents in order to create multiple templatic views, including the generation of multi-modal documents. We believe that our work supports future research in the area of reasoning and long document generation. 
% \sjauhar{It would be nice, if we can make a reasonable argument, to couch this work as a step towards some broader reasoning or content creation goal (many-to-many document synthesis? something else?). It'll strengthen the motivation as well as contextualize the claims we're making here (Rather than say we solved X and we're done, we say we have made strides towards solving Y by doing X)}

\section{Related Work}

There are several areas of related research in NLP that are relevant to the problems of document transformation and evaluation.

Document summarization has been explored in a number of domains, including news~\cite{see-etal-2017-get}, literature~\cite{Scir2023EchoesFA}, law~\cite{Deroy2023HowRA}, and dialogue~\cite{Chen2021DialogSumAR}.
% \sjauhar{If broadening the motivation in the intro to talk more generally about document views of all types, begin by talking about summarization more generally}
In the scientific domain, summarization of scientific papers has taken the form of long form summaries~\cite{chandrasekaran-etal-2020-overview-insights}, abstract generation~\cite{Cohan2015ScientificAS}, conference talks~\cite{Lev2019TalkSummAD}, and query based summaries~\cite{Fok2023QlarifyBS}. These summaries can be either extractive~\cite{Sefid2022SciBERTSUMES} or abstractive~\cite{chandrasekaran-etal-2020-overview-insights}. 

Although the tasks of slide and poster generation have generally been considered separate from scientific summarization, they are related in that both tasks require taking an input article, then organizing and abstracting the information to generate a new document. Past work has developed methods for slide generation from papers~\cite{Hu2015PPSGenLP, Li2021TowardsTS, Hu2015PPSGenLP,Fu2021DOC2PPTAP}, from code~\cite{Wang2023Slide4NCP}, or based on a query~\cite{Sun2021D2SDG}. Poster generation has been explored in the form of content extraction for posters~\cite{Xu2021NeuralCE}, interactive generation~\cite{Wang2015iPosterIP}, or full content generation using graphical models~\cite{Qiang2016LearningTG}. To the best of our knowledge, our work is the first to create a unified method capable of generating a diverse range of templatic views of a source document. %\sjauhar{... and is capable of scaling to others}. 

Large Language Models (LLMs), which are central to our approach, have shown impressive capabilities in a variety of tasks~\cite{Radford2019LanguageMA, Brown2020LanguageMA}. Based on the transformer architecture~\cite{Vaswani2017AttentionIA}, LLMs have shown emergent abilities in tasks such as arithmetic and question answering~\cite{Wei2022EmergentAO}. %Past work has shown that chain of thought prompting can improve performance of downstream tasks~\cite{Wei2022ChainOT, Wang2022SelfConsistencyIC}. \citet{Creo2023PromptingLW} showed that content planning prompting can improve summaries of scientific papers. 
Similar to chain of thought prompting~\cite{Wei2022ChainOT} and content planning prompting~\cite{Wang2022SelfConsistencyIC}, we show that by generating an intermediate representation of an input document can improve performance over simply prompting the model to generate the final document from the original input. 

As past work has tackled generation of templatic views as separate tasks, methods for automatic evaluation of different document types is fragmented. LongSumm, the shared task introduced by~\citet{chandrasekaran-etal-2020-overview-insights}, uses ROUGE to evaluate model performance~\cite{Lin2004ROUGEAP}. \citet{Fu2021DOC2PPTAP} introduced Slide Level ROUGE to evaluate slide generation, a variant that contains a penalty for the number of slides. \citet{Qiang2016LearningTG} used a trained regressor. For summarization, many automatic evaluation metrics have been introduced such as BERTScore~\cite{Zhang2019BERTScoreET}, UniEval~\cite{Zhong2022TowardsAU}, BARTScore~\cite{Yuan2021BARTScoreEG}, BLANC~\cite{Vasilyev2020FillIT}, and MoverScore~\cite{Zhao2019MoverScoreTG}. However, these metrics are intended for a simple input document-summary setup, and do not take into account factors that affect the quality of other types of documents (e.g. structure). Our work is the first to introduce template adaptable evaluation, allowing uniform comparison of performance across template types.

\section{Data}\label{sec:data}

We begin by describing the data used in this paper. There is no existing dataset that includes multiple views of a single document. Instead, we evaluate our unified method, described in \S\ref{sec:method}, on 3 existing datasets: DOC2PPT, LongSumm, and Paper-Poster \cite{Fu2021DOC2PPTAP, chandrasekaran-etal-2020-overview-insights, Qiang2016LearningTG}. These datasets are chosen because they each involve generating a different view of a document. Although our method is not specific to the scientific domain, it is one of the few domains with abundantly available public data of multiple templatic views~\footnote{We acknowledge that scientific writing does have structural regularities that may influence unified document generation. Due to the lack of other available datasets we leave exploration of other domains to future work.}. The three datasets and their associated generation tasks are described below.

\paragraph{Slide Generation.} We use use the DOC2PPT dataset~\cite{Fu2021DOC2PPTAP}, which contains 5.8K scientific papers in Computer Science and their respective slide decks. As \citet{Fu2021DOC2PPTAP} do not release data splits or code, we randomly sample 1K examples from this dataset for evaluation. The slides are provided as an image for each slide. We use the Azure OCR tool to extract the text from each slide\footnote{\scriptsize \url{https://learn.microsoft.com/en-us/azure/ai-services/computer-vision/overview-ocr}}.

\paragraph{Blog Generation.} We use the LongSumm dataset~\cite{chandrasekaran-etal-2020-overview-insights}, which includes blog posts of scientific papers in the Computer Science domain. Since our approach requires no training or supervision, we use the entire training split from Longsumm as our evaluation set. Of the 531 publicly released blog posts in this set, we could only access 505, with the other 26 including broken links or being behind a paywall.

Notably, while Longsumm includes a blind test set of 22 papers, this test set only consists of inputs without their reference outputs, thus making it impossible to compute our custom evaluation metric (see \S\ref{sec:eval-metric}). In the interest of completeness and comparison to prior work, we do, however submit runs from our systems to the leader board and report the results of this blind test set in Appendix~\ref{sec:longsumm-blind}.

%Therefore, we use the larger released train split for our evaluation, since our method does not require any training or supervision. This larger set allows for more robust evaluation than the smaller blind test set. . 

\paragraph{Poster Generation.} We use the Paper-Poster dataset~\cite{Qiang2016LearningTG}, which consists of a dataset of 85 papers in Computer Science and Biology, and their respective scientific posters; two examples containing corrupted PDFs are excluded. Although \citet{Qiang2016LearningTG} release data splits, they do not release code or results for comparison. Given the small size of the dataset, we use it in its entirety for more robust results. While the authors uses the source files to extract the text of posters for evaluation, they only release the PDF formats. To preprocess the reference posters, we found that automatic tools to extract text from documents did not handle the visual layout of posters well, so we manually extracted the text of the posters in this dataset. Note that this process was only done to obtain evaluation scores, and that our unsupervised generation method is capable of creating target documents without the need for reference data.%, or this dataset-specific manual extraction step.
%\sjauhar{You manually transcribed the text from the posters?? I hadn't realized. That is both amazing (for obvious reasons), and concerning (if the reviewers say that this will not scale). If there are assumptions that need to be made for your method to work, those need to be listed out clearly in the Method section.}\isabel{I had to manually extract the poster text for the reference data only, it's not necessary for the generation method, since we automatically extract that with azure. Our evaluation metric does assume you have a reference document, but that's true for all of our comparison metrics.}

% \sjauhar{This contradicts what you say about the poster dataset}\isabel{I manually extracted the text of the posters, the papers were automatically extracted. I edited the above paragraph to hopefully make that more clear}

For all 3 datasets, we use the Azure Document Layout tool to extract the text of the input papers.\footnote{\scriptsize \url{https://learn.microsoft.com/en-us/azure/ai-services/document-intelligence/concept-layout}}

\section{Unified LLM-powered Generation of Templatic Views}\label{sec:method}
The most straightforward way to transform documents between templatic views using LLMs, is to simply prompt the system to generate the target view given the input. However, similar to chain of thought prompting~\cite{Wei2022ChainOT}, we hypothesize that first generating a structured, intermediate representation of an input document and then reasoning over that representation will result in better generations than directly prompting the model. Our goal is to evaluate the capabilites of LLMs to generate long, structured documents, and experiment with how structured prompting can improve performance. We experiment with a simple general two-step process: first generate an intermediate representation, then generate the templatic view. These steps are described in greater detail below, and the process is visualized in Figure~\ref{fig:method}.

\paragraph{Intermediate Representation Generation.} In this work, we set the intermediate representation to be a JSON consisting of a structured layout of the most important parts of the input. We provide the input document to the model along with a template of the representation and prompt it to extract the most important information from the input document, and format it in the given JSON structure. The exact prompts and JSON structure can be found in Appendix~\S\ref{sec:prompt-details}. 
While our experiments use a JSON intermediate representation, note that other formats that provide structure to the input text could be employed (e.g. XML or Markdown). Rather than trying to optimize for the best representation format, our goal is to show that this chain of extraction approach along with structured augmentation to prompts can aid the quality of generations from LLMs. We leave exploration of different formats and other prompt optimization to future work.

\paragraph{Templatic View Generation.} We then feed the generated representation as input back into the LLM, prompting the model to generate the final output document, represented as a LaTeX document. For each templatic view, the prompt to generate the final LaTeX document takes a short description of the desired output, which we refer to as a style parameter. For example, the style parameter for slide generation is as follows: ``Slides should include a title page. Following slides should contain an informative slide title and short, concise bullet points. Longer slides should be broken up into multiple slides.'' The use of style parameters makes our method adaptable to new templatic views; the user only needs to write a short description of the template style. Both the generation of the intermediate representations and the final documents require little to no prompt engineering.  The prompts and style parameters can be found in Appendix~\S\ref{sec:prompt-details}.

\section{Template Adaptable Evaluation}\label{sec:eval-metric}

%As past work has treated the generation of different templatic views as separate tasks, the evaluation of these types has also been developed independently.
%Unfortunately, none of the previously used metrics scale well to other document types or more generally to a unified evaluation of templatic views.
Prior work on document generation has treated the evaluation of different templatic views as separate tasks.
Thus, our goal is to develop a framework of automatic evaluation that is \textit{template adaptable}. This not only allows us to compare performance across diverse datasets, it also removes the requirement of designing and maintaining individual metrics for each template. In order to generalize to multiple templates, we introduce the concept of \textit{panels}. A panel is a unit of organization within a document type, for which the placement and ordering of the panel is important to the overall flow of information in the document.  
% \sjauhar{I am wondering if we should maybe use a different word for this -- say \emph{panels}? Frames have a very definite historical connotation in NLP (frame semantics) and although this shouldn't be confusing because we're dealing with a totally different domain we may want to avoid overloading the term}. \isabel{It's a good point about frames already having another meaning in NLP, but panels feels specific to posters. } \sjauhar{Can you talk about frames more generally in the context of \emph{all} doc types, before focusing on how they would apply to types in this paper?} 

For example, we consider panels to be each slide in a slide deck and each section on a poster. We consider the entirety of a blog post to be a single panel. Although we test our method on the tasks of slide, blog, and poster generation, the concept of panels is not limited to these document types. For example, each post on a social media thread could be considered a panel, or each page on a website.

We aim to unify the evaluation of templatic views by integrating prior metrics into a template adaptable precision-recall framework, which we refer to as Template-Adaptable Evaluation (TAE). TAE is not a new individual metric, but rather an evaluation framework that allows generalization to new templates. For example, TAE can even be used \textit{with} ROUGE to evaluate poster generation. 

% \sjauhar{Are there other aspects of the modularity and extensibility of your eval that are worth mentioning? I think it's important to at least draw attention to the fact that you're defining a \emph{template} for evaluation rather than a single metric. People can instantiate it a variety of different ways, depending on their application (and even train it as a regressor if you add weights) and thus previously used metrics happen to be special cases of your metric template, rather than simply different ones (this makes sense because you're doing something unified vs. disparate)}. 

The general TAE formulation is as follows: 
\begin{equation}
    \begin{split}
    \mathrm{Precision} = Q_P \times O_P \times L \\
    \mathrm{Recall} = Q_R \times O_R \times L
    \end{split}
\end{equation}
\noindent
in which $Q_P$ is the precision measure of quality (\S\ref{sec:qual-measure}), $O_P$ is the precision penalty for order (\S\ref{sec:order-penalty}), and $L$ is the non-reflexive penalty for length (\S\ref{sec:length-penalty}). 
% \sjauhar{In the most general formulation do you want to introduce weights that could place different levels of importance on different terms?}
% \isabel{i could mention weighting but we don't experiment with it at all}\sjauhar{I think that is worth doing, given that the story for the metric is unification but also flexibility}. 
Similarly, $Q_R$ is the recall quality measure and $O_R$ is the recall penalty for order. The precision-recall formulation allows evaluators to decide which measure is most important to them, or calculate an overall F-measure score.

\subsection{Quality Measure}\label{sec:qual-measure}

For the TAE precision score, we calculate the average similarity between the generated panels and their most similar reference panel as follows:%. More specifically:%, we calculate $Q_P$ as follows:

\begin{equation}
    Q_P = \frac{1}{|\tilde{S}|} \sum_{\tilde{S}} \mathrm{max}_{\mathrm{sim}}( S, \tilde{S}_i )
\end{equation}
\noindent
in which $S$ is the set of reference panels and $\tilde{S}$ is the set of generated panels. 
% \sjauhar{Provide a more succinct interpretation first, e.g. ``This is the average similarity between reference frames and their most similar generated frames'' or something like that.} In other words, for each generated frame, we align it to the most similar reference frame, then take the average of the scores. This captures how similar each generated slide is to a reference slide.
For the similarity metric, the user can choose a metric that best matches their use case, such as ROUGE, BERT-Score, or a custom trained regressor~\cite{Lin2004ROUGEAP,Zhang2019BERTScoreET}. For example, a user might choose ROUGE if they want a similarity metric that focuses on exact word overlap, or BERTScore to measure broader semantic similarity.
% \sjauhar{Briefly explain when you would use one vs. the other.}. 
% Additionally, the TAE framework can be used as future work introduces improved similarity metrics.

Similar to precision, the TAE recall score is calculated as the average similarity between the reference panels and their most similar generated panel:

\begin{equation}
    Q_R = \frac{1}{|S|} \sum_{S} \mathrm{max}_{\mathrm{sim}}( \tilde{S}, S_i )
\end{equation}

%In other words, we measure how similar each reference panel is to the generated panels. 

By splitting the evaluation of quality into precision and recall, we can evaluate both the content of the slides that were generated as well as the coverage of this content against some reference.

\subsection{Order Penalty}\label{sec:order-penalty}

% \sjauhar{How about wording it as follows:}

% \subsubsection{Alternative Version}

\begin{figure}[t!]
    \centering
    \includegraphics[width=.8\columnwidth]{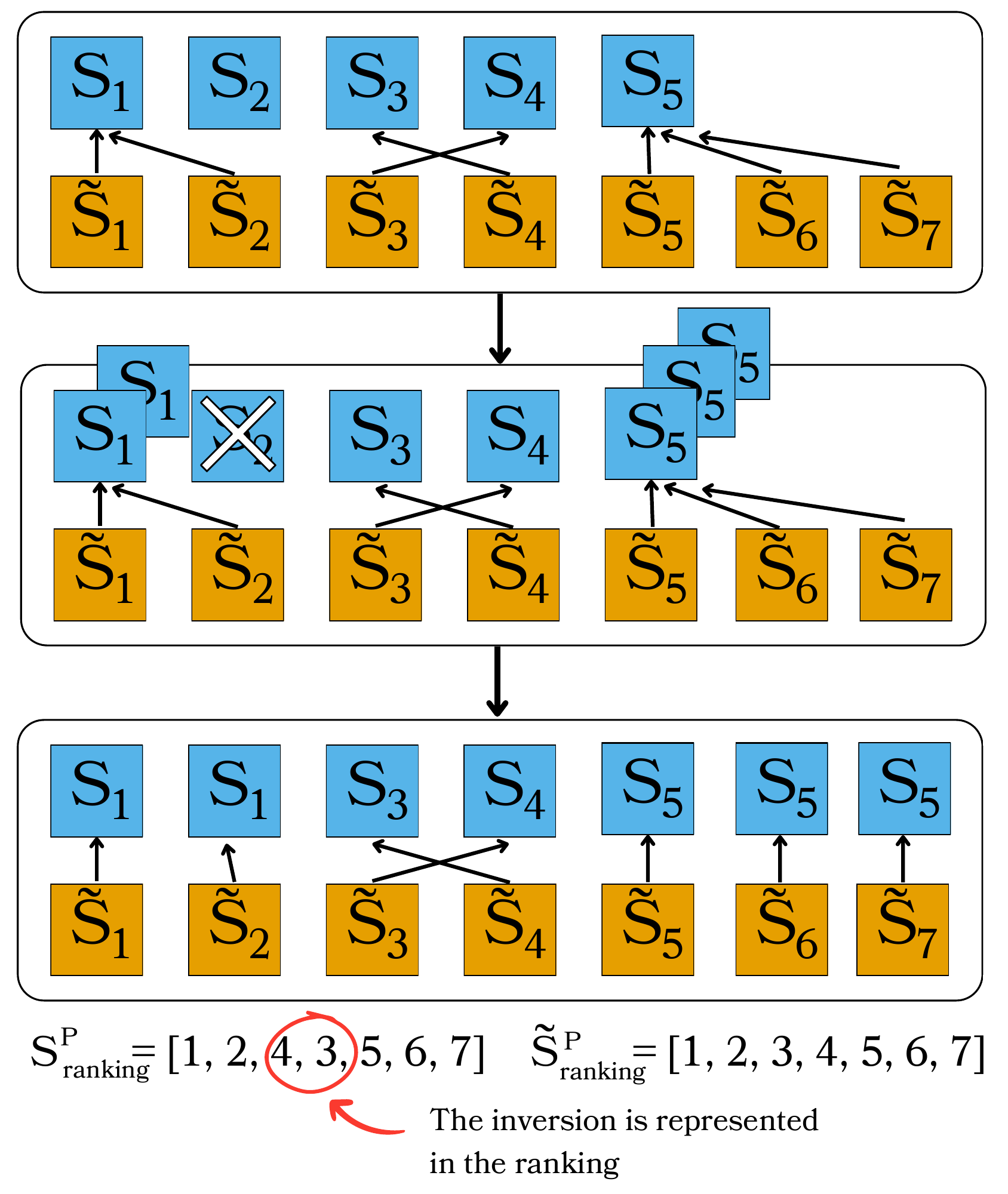}

    \caption{\small Example of the process of obtaining the rankings for the precision ordering penalty. We first use the similarity measure to map each generated panel to its most similar reference document. This mapping is used to calculate the precision quality score $Q_P$. We then use the mappings to create a one-to-one alignment from the generated to the reference panels, which we use to calculate the precision ordering penalty ($O_P$). By creating a one-to-one alignment, we are able to represent inversions in the ordering.
    This process is reflexive, and panels not accounted for in the precision ordering penalty are accounted for in the recall ordering penalty.
    } 
    \label{fig:ordering-penalty}
\end{figure}

Broadly, the goal of the ordering penalty is to measure the similarity of the \emph{order} of information in reference and generated panels, independent of other factors. Unfortunately, because the cardinality of panels in the two outputs is not necessarily the same, a direct one-to-one mapping to compare ordering is not feasible. Instead, a panel in one set can align to multiple references in the other, or none at all -- as depicted in ~Figure~\ref{fig:ordering-penalty}. Intuitively, our solution is to virtually replicate (resp. drop) panels that have multiple (resp. zero) alignments in the reference set so that a one-to-one mapping of ordering, can in fact be computed.
% \sjauhar{I am being lazy here and assuming the ordering for precision; you may need to polish this a bit to explain how it will work in reverse for recall}

Formally, assume $S$ and $\tilde{S}$  are sequences of reference and generated panels respectively. We use the maximum similarity scores calculated in \S\ref{sec:qual-measure} to align the panels across sets.

For the precision ordering penalty, we define the following operation $\lambda_P(s)=\sum_{\tilde{s}}\delta_P(s,\tilde{s})$, where
\[
    \delta_P(s,\tilde{s})= 
\begin{cases}
    1, & \text{iff } s \rightarrow \tilde{s}\\
    0, & \text{otherwise}
\end{cases}
\]

\noindent Intuitively, this captures the cardinality of the alignment of a panel in $S$ with panels in $\tilde{S}$. Then, using this operation we can replace every $s \in S$ with $\lambda_P(s)$ copies, leading to an identical cardinality for both $S$ and $\tilde{S}$, and subsequent one-to-one mapping between their corresponding panels.

%With this one-to-one mapping in place we can then measure the misalignments between the two sets of ordered panels to assign a penalty scores.
Then, to operationalize a penalty score for the two sets of ordered panels we associate them with ranks in both sets and use a rank correlation metric to compute the degree of agreement. Specifically, rank assignment is done as follows: panels in $\tilde{S}$ are simply assigned ranks in order of appearance 1 through N -- we call this $\tilde{S}_{ranking}^P$; meanwhile panels in $S$ are assigned the identical rank to their one-to-one aligned panel in $\tilde{S}$ and $\tilde{S}_{ranking}^P$ -- we refer to these rankings as $S_{ranking}^P$.  An example of this process can be found in Figure~\ref{fig:ordering-penalty}. The final ordering penalty is computed using Spearman's rank correlation~\cite{5548399}:
\vspace{-0.5mm}
\begin{equation}
    O_P = \frac{\mathrm{Spearman}(S_{ranking}^P, \tilde{S}_{ranking}^P) + 1}{2}
\end{equation}

\noindent where we perform a linear transformation to map the original range of the correlation coefficient [-1, 1] to the desired range [0, 1].

Similarly, for the recall ordering penalty, we map the reference panels to the generated panels, calculated as $\lambda_R(s)=\sum_{s}\delta_R(\tilde{s}, s)$. $O_R$ is calculated similar to $O_P$, using the recall rankings.%, where

\subsection{Length Penalty}\label{sec:length-penalty}

Finally, we compute a length penalty for both the recall and precision scores. Similar to \citet{Fu2021DOC2PPTAP}, this is done as follows:
\begin{equation}
    L = e^{\frac{-\mathrm{abs}(|S| - |\tilde{S}|)}{|S|}}
\end{equation}

\noindent We chose to keep $L$ non-reflexive, because in the reverse case -- as $|\tilde{S}| \rightarrow \infty$, $L \rightarrow 1$ -- the metric could be cheated by over-generating.

%If $L$ were reflexive, the denominator in the reverse case would be $|\tilde{S}|$, meaning that as $|\tilde{S}| \rightarrow \infty$, $L \rightarrow 1$. In other words, the length penalty could be cheated by over-generating. Therefore, unlike the quality measure and ordering penalty, we chose not to make the length penalty reflexive. 

\section{Results}
As mentioned in \S\ref{sec:data}, past work on Doc2PPT and Paper-Posters do not release code, making it difficult to do a direct comparison. They also do not report any baselines to compare against. Meanwhile, Longsumm's blind test does not allow us to compute our custom metric, although we do report the leaderboard results in Appendix~\ref{sec:longsumm-blind}. Notably, with almost no prompt engineering our LLM-based system places second on this leaderboard. We argue that for the investigation in this paper, direct comparison to prior non-LLM baselines is not only unfair to those approaches, but not particularly insightful.
Therefore, similar to~\citet{Wei2022ChainOT}, we focus on variants of our LLM-based method and treat them as baselines. Example outputs of each template type can be found in Appendix~\ref{sec:example-outputs}.

We conduct experiments with the following settings:
\begin{enumerate*}[label={(\arabic*)}]
    \item No Representation -- this is the default setting of going directly from the source document to the target document. We skip the intermediate generation step, passing the full paper as input. We experiment both with and without the style parameters. 
    \item Own Representation -- we do not pass a JSON structure to the intermediate generation step, and allow the model to choose its own structure.
    \item Text Representation -- we extract the text from the intermediate representation, discarding the JSON structure.
    \item JSON Representation -- this is the full JSON structure for the intermediate generation step. We experiment both with and without the style parameters.
\end{enumerate*}

 %\isabel{Should we run on the blind set and submit it to their leaderboard? It's only 22 papers.} \sjauhar{Might be worth doing for the ARR submission.}
% \sjauhar{This feels more appropriate when talking about the experimental results; maybe move?} 
We use \texttt{gpt35-16k} in our main set of experiments. We truncate text that is too long for the input window and use a temperature of 0.0 as standard.\footnote{A detailed evaluation of the temperature hyper-parameter is included in Appendix \S\ref{sec:temp-exp}}
\begin{table}[!t]
\centering
\small
\begin{tabular}{ccc|cccc}
 &  &  & \multicolumn{4}{c}{Similarity Measure} \\
 & Rep. & Style & R-L & M & B & BERTS \\ \hline
\multirow{6}{*}{\rotatebox[origin=c]{90}{Slides}} & None & $\times$ & 5.0 & 6.4 & 0.3 & 31.6 \\
 & None & $\checkmark$ & 5.1 & 6.0 & 0.4 & 31.7 \\
 & Own & $\checkmark$ & 6.5 & 7.1 & 1.2 & 36.1 \\
& Text & $\checkmark$ & 7.3 & 8.0 & 1.4 & 36.4 \\
 & JSON & $\times$ & 4.2 & 6.0 & 0.3 & 31.4 \\ 
 & JSON & $\checkmark$ & \textbf{7.4} & \textbf{8.4} & \textbf{1.5} & \textbf{36.9} \\ 
 \hdashline[0.5pt/3pt]
 
\multirow{6}{*}{\rotatebox[origin=c]{90}{Blogs}} & None & $\times$ & 26.6 & 19.6 & 3.0 & 82.5 \\
 & None & $\checkmark$ & 25.1 & 17.7 & 2.3 & \textbf{82.8} \\
 & Own & $\checkmark$ & 23.9 & 19.2 & 2.3 & 82.2 \\
& Text & $\checkmark$ & 25.4 & 19.3 & 2.5 & 82.5 \\
 & JSON & $\times$ & \textbf{28.3} & \textbf{25.3} & \textbf{5.0} & 82.3 \\ 
 & JSON & $\checkmark$ & 25.4 & 19.6 & 2.8 & 82.4 \\ \hdashline[0.5pt/3pt]
 
\multirow{6}{*}{\rotatebox[origin=c]{90}{Posters}} & None & $\times$ & 8.1 & 10.3 & 1.0 & 35.6 \\
 & None & $\checkmark$ & 10.1 & 11.6 & 1.9 & 39.5 \\
 & Own & $\checkmark$ & 12.8 & 12.6 & 2.9 & 52.8 \\
& Text & $\checkmark$ & 11.3 & 11.7 & 2.1 & 45.9 \\
 & JSON & $\times$ & 14.2 & \textbf{16.8} & 4.0 & 52.8 \\ 
 & JSON & $\checkmark$ & \textbf{15.5} & 14.5 & \textbf{15.3 }& \textbf{53.3} \\ %\hdashline[0.5pt/3pt]
\end{tabular}

\caption{\small Evaluation results using GPT3.5 (\texttt{gpt35-16k}). For each template, we experiment with different representations (Rep) and whether or not we include the style parameters (Style). We report the TAE F1 scores as calculated in \S\ref{sec:eval-metric}, using ROUGE-L (R-L), METEOR (M), BLEU (B), and BERTScore (BERTS) as the similarity metrics.}
\label{tbl:auto-results}
\end{table}
\subsection{Results of automatic evaluation}\label{sec:auto-results}

In Table~\ref{tbl:auto-results}, we report the TAE F1 scores as described in \S\ref{sec:eval-metric}, using ROUGE-L~\cite{Lin2004ROUGEAP}, METEOR~\cite{Banerjee2005METEORAA}, BLEU~\cite{Papineni2002BleuAM}, and BERTScore~\cite{Zhang2019BERTScoreET} for the similarity measure. As seen in the results, by most measures, generating a JSON intermediate representation yields the best performance. 

We see that using the text representation generally degrades the performance over providing the structured JSON representation, indicating that structure is important for downstream performance in addition to abstractive filtering of information. Additionally, the text representation performs better than skipping the intermediate step altogether for both the poster and slide generation task, but not the blog generation task. This is likely because posters and slides have more inherent structure than blog posts, which can be relatively free-form.

Finally, we see that allowing the model to choose its own representation format degrades performance over providing our pre-defined JSON structure. However, we see that in most cases, providing a representation generated without a JSON structure still performs better than skipping the intermediate generation step altogether (while maintaining the same style parameter setting). This indicates that even without a pre-defined structure, the intermediate step is still valuable for performance. 

% \usepackage{multirow}
\begin{table}[]
\centering
\small
\begin{tabular}{ccc|cccc}
 &  &  & \multicolumn{4}{c}{Similarity Measure} \\
 & Model & Rep. & R-L & M & B & BERTS \\ \hline
\multirow{6}{*}{\rotatebox[origin=c]{90}{Slides}} & \multirow{2}{*}{MS} & $\times$ & 0.6 & 0.4 & 0.0 & 28.6 \\
 &  & $\checkmark$ & \textbf{4.4} & \textbf{4.6} & \textbf{0.4} & \textbf{30.5} \\ \cdashline{2-7}[0.5pt/3pt]
 & \multirow{2}{*}{MX} & $\times$ & 4.8 & 7.3 & 0.5 & 31.8 \\
 &  & $\checkmark$ & \textbf{6.7} & \textbf{7.9} & \textbf{1.0} & \textbf{34.0} \\ \cdashline{2-7}[0.5pt/3pt]
 & \multicolumn{1}{c}{\multirow{2}{*}{GPT4}} & \multicolumn{1}{c|}{$\times$} & \multicolumn{1}{c}{8.3} & \multicolumn{1}{c}{\textbf{9.6}} & \multicolumn{1}{c}{1.7} & \multicolumn{1}{c}{36.2} \\
 & \multicolumn{1}{c}{} & \multicolumn{1}{c|}{$\checkmark$} & \multicolumn{1}{c}{\textbf{8.4}} & \multicolumn{1}{c}{9.1} & \multicolumn{1}{c}{\textbf{2.0}} & \multicolumn{1}{c}{\textbf{38.1}} \\  \hdashline[0.5pt/3pt]
\multirow{6}{*}{\rotatebox[origin=c]{90}{Blog}} & \multirow{2}{*}{MS} & $\times$ & 2.7 & 1.7 & 0.1 & 73.5 \\
 &  & $\checkmark$ & \textbf{21.7} & \textbf{16.2} & \textbf{1.7} & \textbf{81.3} \\ \cdashline{2-7}[0.5pt/3pt]
 & \multirow{2}{*}{MX} & $\times$ & 22.8 & 15.7 & 2.1 & \textbf{82.6} \\
 &  & $\checkmark$ & \textbf{25.6} & \textbf{20.5} & \textbf{3.2} & 82.5 \\ \cdashline{2-7}[0.5pt/3pt]
 & \multicolumn{1}{c}{\multirow{2}{*}{GPT4}} & \multicolumn{1}{c|}{$\times$} & \multicolumn{1}{c}{25.7} & \multicolumn{1}{c}{19.9} & \multicolumn{1}{c}{2.6} & \multicolumn{1}{c}{\textbf{82.8}} \\
 & \multicolumn{1}{c}{} & \multicolumn{1}{c|}{$\checkmark$} & \multicolumn{1}{c}{\textbf{25.8}} & \multicolumn{1}{c}{\textbf{20.2}} & \multicolumn{1}{c}{\textbf{3.1}} & \multicolumn{1}{c}{82.7} \\  \hdashline[0.5pt/3pt]
\multirow{6}{*}{\rotatebox[origin=c]{90}{Poster}} & \multirow{2}{*}{MS} & $\times$ & 3.2 & 1.8 & 0.2 & 32.2 \\
 &  & $\checkmark$ & \textbf{6.0} & \textbf{6.5} & \textbf{1.2} & \textbf{38.1} \\ \cdashline{2-7}[0.5pt/3pt]
 & \multirow{2}{*}{MX} & $\times$ & \textbf{10.5} & \textbf{11.4} & \textbf{1.7} & 40.9 \\
 &  & $\checkmark$ & 10.4 & 11.0 & 1.5 & \textbf{50.7} \\ \cdashline{2-7}[0.5pt/3pt]
 & \multicolumn{1}{c}{\multirow{2}{*}{GPT4}} & \multicolumn{1}{c|}{$\times$} & \multicolumn{1}{c}{\textbf{16.4}} & \multicolumn{1}{c}{\textbf{18.2}} & \multicolumn{1}{c}{\textbf{4.5}} & \multicolumn{1}{c}{\textbf{59.8}} \\
 & \multicolumn{1}{c}{} & \multicolumn{1}{c|}{$\checkmark$} & \multicolumn{1}{c}{14.6} & \multicolumn{1}{c}{15.3} & \multicolumn{1}{c}{3.7} & \multicolumn{1}{c}{57.2}
\end{tabular}
\caption{\small  TAE F1 scores using Mistral-7b (MS), Mixtral (MX), and GPT4. We use ROUGE-L (R-L), METEOR (M), BLEU (B) and BERTScore (BERTS) as our similarity measures. For each template, we compare a JSON representation versus skipping the intermediate generation step (Rep), maintaining the same style parameters in both settings.}\label{tbl:other-model-results}
\end{table}
% \subsubsection{Test a higher performing model}
\paragraph{Experiments with additional models.} We conduct a subset of our experiments on Mistral-7B~\cite{Jiang2023Mistral7}, Mixtral~\cite{Jiang2024MixtralOE}, and GPT4 ($\texttt{gpt4-32k})$, comparing the JSON representation to skipping the intermediate step.  We maintain the same style parameters in both settings. In Table~\ref{tbl:other-model-results}, we can see that by most measures, the documents generated with the intermediate representation score higher than the documents generated without, particularly for blog posts and slides. The difference in performance is larger for Mistral than Mixtral and GPT4, indicating that our method particularly improves the performance of smaller models. Smaller models are generally cheaper, less resource intensive, and faster, but often operate at the cost of performance. The results indicate that for applications that are sensitive to cost or latency, this trade-off can be mitigated with a structured intermediate representation. The only experiment in which the documents generated with the representation do not strictly score higher on most measures is the posters generated with Mixtral and GPT4. Upon closer inspection, the references in this dataset are very verbose, averaging 391 tokens. Our method produces generally less verbose posters, averaging 265 total tokens compared to 345 tokens produced by the baseline. We hypothesize that by editing the style parameters to include information about verbosity and length, we can improve performance on posters in the future.

\begin{figure}
    \centering
    \includegraphics[width=.8\columnwidth]{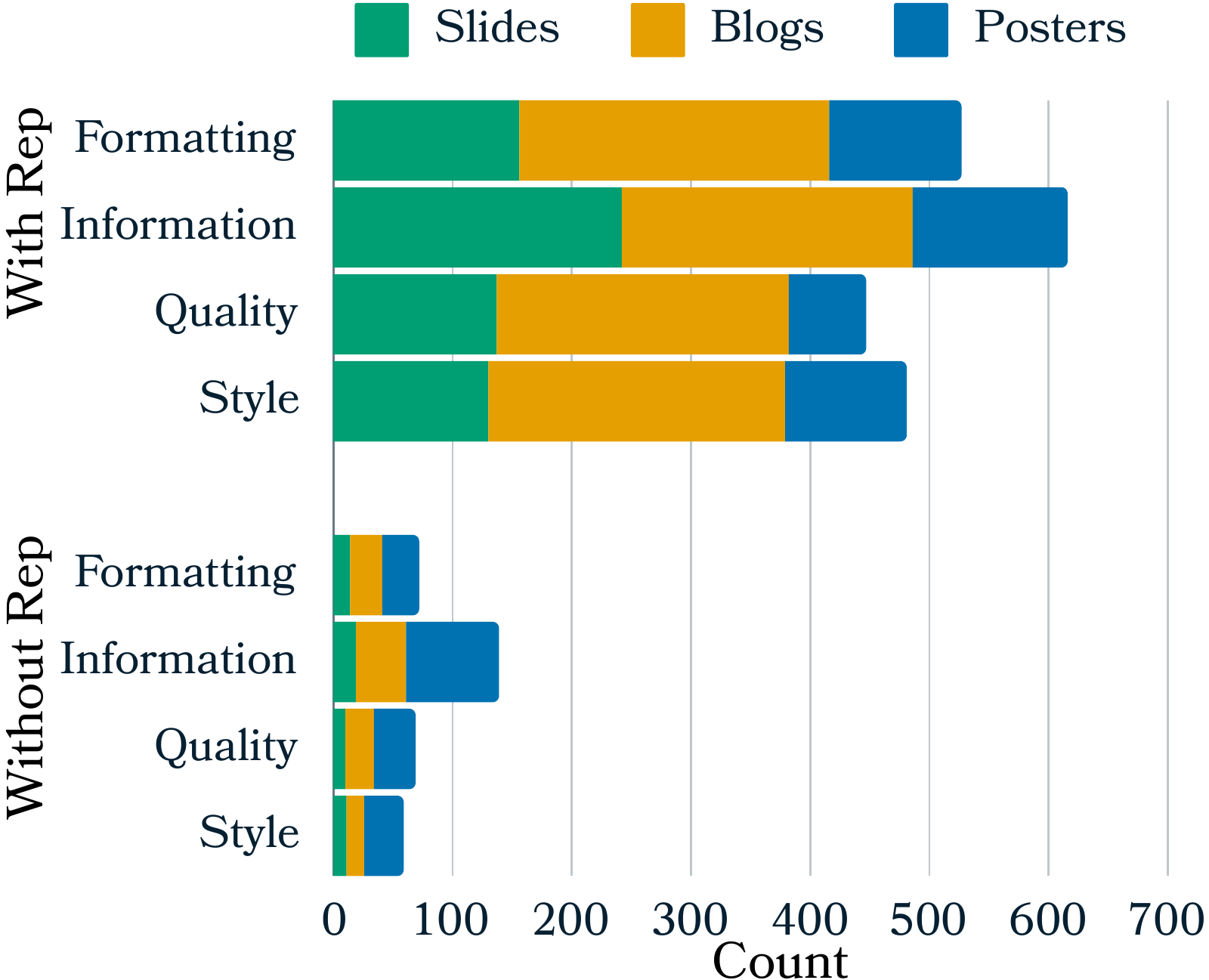}\vspace*{-1mm}
    \caption{\small Reasons annotators preferred each document. While annotators largely preferred documents generated with an intermediate representation, the most common reasons for preference are better formatting and information content. We exclude the ``Other'' count as it was only selected once.}
    \label{fig:annotation-reasons}\vspace*{-3mm}
\end{figure}

\subsection{Human evaluation}
After showing that LLMs benefit from intermediate structured representations in document transformations, we investigate whether our proposed evaluation framework aligns better with human judgment than previously proposed metrics. We sample 100 documents each from DOC2PPT and LongSumm, and use the entirety of the Paper-Poster dataset in this study. We present annotators with 2 versions of each document, one generated with the intermediate representation and one without. Both versions use $\texttt{gpt4-32k}$, as the best performing model. 

The annotators are provided with the original paper and the intended document type (blog, slide deck, or poster), and are asked the following questions:
\begin{enumerate*}[label={(\arabic*)}]
    \item Which document do you prefer?
    %\vspace{-2mm}
    \item  On a scale of 1-3, to what degree do you prefer your selection?
    %\vspace{-2mm}
    \item Why do you prefer your selection? 
\end{enumerate*}
For question 3, annotators are also provided with a multi-select checklist of reasons for their preference: (1) quality of the content, (2) formatting, (3) document style matching the intended document types, (4) information represented in the document, and (5) other (along with a free text box). The full instructions, including the reasons provided and examples, can be found in Appendix~\S\ref{sec:annotation-instructions}.

If the models do not produce LaTeX and instead produce only text, we wrap the text with $\texttt{\textbackslash begin\{document\}}$ and $\texttt{\textbackslash end\{document\}}$. We force  the compilation of the outputs with the command: $\texttt{pdflatex -{}-interaction=nonstopmode}$ $\texttt{<filename.tex>}$. Occasionally, this forced compilation leads to oddly formatted documents, but we consider this to be a part of the performance of the method and present the documents with no further changes. Each document is annotated by 3 different annotators. We employed 4 annotators from India, sourced via a third-party agency, to carry out the human evaluation of our task based on a guideline document containing task-specific instructions, guidance, and annotated examples. They were compensated at a rate of \$11.98 USD per hour for the total time spent working on the task, including a training round of annotation.

\vspace{-1mm}
\paragraph{Which method do humans prefer?} The documents generated with an intermediate representation were preferred by 82\%, based on majority vote (71\% unanimously). The annotator agreement score was 0.51 with Krippendorff's alpha, indicating that while this is a subjective and specialized task, even non-expert annotators agree to a moderate degree. 
A visualization of the reasons the annotators preferred their selection can be found in Figure~\ref{fig:annotation-reasons}. 
It can be seen that while annotators largely preferred the documents generated with an intermediate representation, the most common reasons for preference are better formatting and better information content. This indicates that the structure provided by the intermediate representation makes it easier for the model to format the final document well. Additionally, the intermediate representation only includes the most salient information from the original text, resulting in higher quality of information content. Finally, we see a fairly even distribution across different templatic views for the reasons of preference, indicating humans prefer the documents generated with the intermediate representation across different document types.
% \sjauhar{Can you say something about why a structured intermediate representation may lead to these artifacts?}

% \input{tables/human_eval_correlation}

\paragraph{Which metric correlates better with humans?} We test whether our metric, as described in \S\ref{sec:eval-metric}, correlates better with human preference compared to prior evaluation metrics in the literature.
%We use the human preference judgements to test if our metric, as described in \S\ref{sec:eval-metric}, correlates more with human preference compared to prior evaluation metrics in the literature. %\sjauhar{This may be a bit misleading, since you're actually comparing correlation \emph{against} other metrics. I think it's important to contextualize this by explaining that what we want to show is that our metric should be preferred to other metrics vs. just being highly correlated with human judgement (which it is not)}. 
For each annotation, given the degree of the preference $d$ (Appendix~\S\ref{sec:annotation-instructions} Q. 2) we convert value to a score $P(d) \rightarrow [1,2,3]$ if $d$ is slight, moderate, or strong, respectively. 
% \[
%     P(d) = 
% \begin{cases}
%     3, & d=\textrm{strong} \\
%     2, & d=\textrm{moderate} \\
%     1, & d=\textrm{slight}
% \end{cases}
% \]
If the annotator prefers the document generated without an intermediate representation, we take $-P(d)$ instead. This allows us to measure if the metric captures directionality of preference along with degree. in parallel, we compute the automatic score $m$ for each metric, then calculate
$S =m(\textrm{with rep}) - m(\textrm{skip rep})$
where $m$ is the metric we are evaluating (e.g ROUGE). If a human annotator prefers a document generated without the intermediate step, we'd expect a good metric to assign a higher score to that document as well, resulting in both $S$ and $P(d)$ being negative (and positive in the opposite case). % Conversely, we'd expect instances where the opposite were true to result in positive scores for both $S$ and $P$.
Using this intuition we assign an affinity score of a metric with respect to human evaluation as the Pearson correlation~\cite{freedman2007statistics} of $S$ and $P(d)$.
% \sjauhar{This whole bit about computing the correlation score is pretty confusing (esp. I'd imagine for a first-time reader); consider structuring more clearly and with better notation.} 

Since prior metrics are not designed to account for the structure of documents, we compute them by extracting only the text of both the generated and reference documents. The correlations with human judgement for each metric to its respective TAE variants can be found in Table~\ref{tbl:correlation}. As we can see from the results, evaluations using our template adaptable framework correlate more highly with human judgement, except in the case of BERTScore. In the latter case the results are not statistically significant, and we hypothesize that the open-domain nature of BERT embeddings are poorly suited to represent the semantic similarity of scientific text.
%However, BERTScore is the only metric in which the correlation is not statistically significant, as it is likely simply a poor metric for this task. We hypothesize that BERTScore does not perform well on scientific data, since it is trained primarily on out of domain text.

\begin{table}
\small
\centering
\begin{tabular}{r|l}
Metric & PearsonR \\ \hline
% ROUGE-1 & 14.9 \\ 
% TAE ROUGE-1 & \textbf{21.4} \\ 
% \hdashline[0.5pt/3pt]
ROUGE-L & 14.5 \\ 
TAE ROUGE-L & \textbf{19.7} \\
\hdashline[0.5pt/3pt]
METEOR & 24.6 \\ 
TAE METEOR & \textbf{25.2} \\
\hdashline[0.5pt/3pt]
BLEU & 13.6 \\ 
TAE BLEU & \textbf{13.8} \\
\hdashline[0.5pt/3pt]
BERTScore & \textbf{10.6*} \\
TAE BERTScore & 5.4* \\ 
\end{tabular}
\caption{\small Correlation of evaluation metrics with human judgement. We compare each metric computed using the TAE framework versus the standard computation. *Indicates the correlation is not statistically significant ($p>0.01$).}
\label{tbl:correlation}\vspace*{-2mm}

\end{table}

\section{Conclusion}

In many domains, people choose to disseminate information across different modalities and formats for better communication to broader audiences. We proposed a unified view of document transformation and evaluation. We showed that LLMs are capable of templatic document generation with minimal supervision, and that a structured, intermediate representation can improve performance, particularly for smaller models. We also introduced a flexible precision-recall framework for automatic evaluation that easily integrates existing evaluation metrics into a unified system and allows for comparison across diverse datasets without additional task specific metric design. Finally, we conducted a human evaluation and showed that annotators prefer the documents generated using the intermediate representation 82\% of the time and that our evaluation framework correlates better with human preference than standard evaluation metrics.% We see this  work as an important step towards the larger goal of generalized document generation. 
%We additionally showed that our framework of evaluation correlates more highly with human preference than standard evaluation metrics.  

% \sjauhar{Consider breaking para here for future work.} The use of templatic views of documents exists in other domains (e.g. creating a cover letter and personal website based on a resume), and in future work we will consider other domains of documents. We position this paper as a step towards the larger goal of many-to-many, multi-modal document template generation. \isabel{I don't like this last sentence, it's not mic drop enough} \sjauhar{Lol, I agree. The last para you had written for the intro (which I commented out) would fit quite well here; consider moving it and adapting into a separate paragraph focused on future work.}

% \sjauhar{Re-iterate the broader impact of this work succintly from the intro.}

\section{Limitations}
% The use of templatic views of documents is relevant for many domains (e.g. creating a cover letter and personal website based on a resume). In future work, we plan to tackle other domains and document types, as well as more complex scenarios, such as many-to-many document generation. We see this  work as an important step towards the larger goal of generalized document generation. 

Although our methods are not domain specific, we only evaluated them in the scientific domain, due to the availability of public data. Additionally, our framework is limited to textual content. In future work we plan to explore the application of our unified framework for generation and evaluation on document views in other domains, as well as incorporating multi-modal models and content generation. Finally, it is possible that some of our test data has leaked into the training data of the models with which we experimented. This limitation is not unique to our work and exists for our baselines in addition to our methods. 

\section{Ethics}
The potential risks of our work are similar to those of other work in downstream applications of LLMs. LLM generated documents can potential generate copy-righted material~\cite{Carlini2020ExtractingTD}, personally-identifiable information~\cite{Lukas2023AnalyzingLO}, or factually incorrect text~\cite{Manakul2023SelfCheckGPTZB}. The use of LLMs to generate documents may violate some academic dishonesty policies~\cite{Zdravkova2023IntegrationOL}. Our system is intended to be used in collaboration with human writers. Users should edit the generations, checking for factual inconsistencies and other potential errors. Our work is intended to save users time that might be spent repeating information across multiple documents, so they can focus on content creation. Therefore, we believe the benefits of our work outweigh the potential risks.

% Additionally, we leave incorporating multi-modal models \sjauhar{This feels like something you could mention in Limitations. Might also be nice to say a few words of the importance of multi-modality in view generation.} into the generation of templatic views for future work.
% \isabel{Do we want to include a limitations and/or ethics section for the preprint?}

% \section*{Ethics}

% \section*{Acknowledgements}
% \todoit{Add acknowledgements for the preprint if needed}

\bibliography{anthology,custom}

\begin{thebibliography}{40}
\expandafter\ifx\csname natexlab\endcsname\relax\def\natexlab#1{#1}\fi

\bibitem[{Banerjee and Lavie(2005)}]{Banerjee2005METEORAA}
Satanjeev Banerjee and Alon Lavie. 2005.
\newblock \href {https://api.semanticscholar.org/CorpusID:7164502} {Meteor: An automatic metric for mt evaluation with improved correlation with human judgments}.
\newblock In \emph{IEEvaluation@ACL}.

\bibitem[{Bornmann and Mutz(2014)}]{Bornmann2014GrowthRO}
Lutz Bornmann and R{\"u}diger Mutz. 2014.
\newblock \href {https://api.semanticscholar.org/CorpusID:7826703} {Growth rates of modern science: A bibliometric analysis based on the number of publications and cited references}.
\newblock \emph{Journal of the Association for Information Science and Technology}, 66.

\bibitem[{Brown et~al.(2020)Brown, Mann, Ryder, Subbiah, Kaplan, Dhariwal, Neelakantan, Shyam, Sastry, Askell, Agarwal, Herbert-Voss, Krueger, Henighan, Child, Ramesh, Ziegler, Wu, Winter, Hesse, Chen, Sigler, Litwin, Gray, Chess, Clark, Berner, McCandlish, Radford, Sutskever, and Amodei}]{Brown2020LanguageMA}
Tom Brown, Benjamin Mann, Nick Ryder, Melanie Subbiah, Jared~D Kaplan, Prafulla Dhariwal, Arvind Neelakantan, Pranav Shyam, Girish Sastry, Amanda Askell, Sandhini Agarwal, Ariel Herbert-Voss, Gretchen Krueger, Tom Henighan, Rewon Child, Aditya Ramesh, Daniel Ziegler, Jeffrey Wu, Clemens Winter, Chris Hesse, Mark Chen, Eric Sigler, Mateusz Litwin, Scott Gray, Benjamin Chess, Jack Clark, Christopher Berner, Sam McCandlish, Alec Radford, Ilya Sutskever, and Dario Amodei. 2020.
\newblock \href {https://proceedings.neurips.cc/paper_files/paper/2020/file/1457c0d6bfcb4967418bfb8ac142f64a-Paper.pdf} {Language models are few-shot learners}.
\newblock In \emph{Advances in Neural Information Processing Systems}, volume~33, pages 1877--1901. Curran Associates, Inc.

\bibitem[{Carlini et~al.(2020)Carlini, Tram{\`e}r, Wallace, Jagielski, Herbert-Voss, Lee, Roberts, Brown, Song, Erlingsson, Oprea, and Raffel}]{Carlini2020ExtractingTD}
Nicholas Carlini, Florian Tram{\`e}r, Eric Wallace, Matthew Jagielski, Ariel Herbert-Voss, Katherine Lee, Adam Roberts, Tom~B. Brown, Dawn~Xiaodong Song, {\'U}lfar Erlingsson, Alina Oprea, and Colin Raffel. 2020.
\newblock \href {https://api.semanticscholar.org/CorpusID:229156229} {Extracting training data from large language models}.
\newblock In \emph{USENIX Security Symposium}.

\bibitem[{Chandrasekaran et~al.(2020)Chandrasekaran, Feigenblat, Hovy, Ravichander, Shmueli-Scheuer, and de~Waard}]{chandrasekaran-etal-2020-overview-insights}
Muthu~Kumar Chandrasekaran, Guy Feigenblat, Eduard Hovy, Abhilasha Ravichander, Michal Shmueli-Scheuer, and Anita de~Waard. 2020.
\newblock \href {https://doi.org/10.18653/v1/2020.sdp-1.24} {Overview and insights from the shared tasks at scholarly document processing 2020: {CL}-{S}ci{S}umm, {L}ay{S}umm and {L}ong{S}umm}.
\newblock In \emph{Proceedings of the First Workshop on Scholarly Document Processing}, pages 214--224, Online. Association for Computational Linguistics.

\bibitem[{Chen et~al.(2021)Chen, Liu, Chen, and Zhang}]{Chen2021DialogSumAR}
Yulong Chen, Yang Liu, Liang Chen, and Yue Zhang. 2021.
\newblock \href {https://doi.org/10.18653/v1/2021.findings-acl.449} {{D}ialog{S}um: {A} real-life scenario dialogue summarization dataset}.
\newblock In \emph{Findings of the Association for Computational Linguistics: ACL-IJCNLP 2021}, pages 5062--5074, Online. Association for Computational Linguistics.

\bibitem[{Cohan and Goharian(2015)}]{Cohan2015ScientificAS}
Arman Cohan and Nazli Goharian. 2015.
\newblock \href {https://api.semanticscholar.org/CorpusID:5523604} {Scientific article summarization using citation-context and article’s discourse structure}.
\newblock In \emph{Conference on Empirical Methods in Natural Language Processing}.

\bibitem[{Deroy et~al.(2023)Deroy, Ghosh, and Ghosh}]{Deroy2023HowRA}
Aniket Deroy, Kripabandhu Ghosh, and Saptarshi Ghosh. 2023.
\newblock \href {https://api.semanticscholar.org/CorpusID:259064225} {How ready are pre-trained abstractive models and llms for legal case judgement summarization?}
\newblock \emph{ArXiv}, abs/2306.01248.

\bibitem[{Fok et~al.(2023)Fok, Chang, August, Zhang, and Weld}]{Fok2023QlarifyBS}
Raymond Fok, Joseph~Chee Chang, Tal August, Amy~X. Zhang, and Daniel~S. Weld. 2023.
\newblock \href {https://api.semanticscholar.org/CorpusID:263835343} {Qlarify: Bridging scholarly abstracts and papers with recursively expandable summaries}.
\newblock \emph{ArXiv}, abs/2310.07581.

\bibitem[{Freedman et~al.(2007)Freedman, Pisani, and Purves}]{freedman2007statistics}
David Freedman, Robert Pisani, and Roger Purves. 2007.
\newblock Statistics (international student edition).
\newblock \emph{Pisani, R. Purves, 4th edn. WW Norton \& Company, New York}.

\bibitem[{Fu et~al.(2021)Fu, Wang, McDuff, and Song}]{Fu2021DOC2PPTAP}
Tsu-Jui Fu, William~Yang Wang, Daniel~J. McDuff, and Yale Song. 2021.
\newblock \href {https://api.semanticscholar.org/CorpusID:231719374} {Doc2ppt: Automatic presentation slides generation from scientific documents}.
\newblock In \emph{AAAI Conference on Artificial Intelligence}.

\bibitem[{Hu and Wan(2015)}]{Hu2015PPSGenLP}
Yue Hu and Xiaojun Wan. 2015.
\newblock \href {https://api.semanticscholar.org/CorpusID:15977803} {Ppsgen: Learning-based presentation slides generation for academic papers}.
\newblock \emph{IEEE Transactions on Knowledge and Data Engineering}, 27:1085--1097.

\bibitem[{Jiang et~al.(2024)Jiang, Sablayrolles, Roux, Mensch, Savary, Bamford, Chaplot, de~Las~Casas, Hanna, Bressand, Lengyel, Bour, Lample, Lavaud, Saulnier, Lachaux, Stock, Subramanian, Yang, Antoniak, Scao, Gervet, Lavril, Wang, Lacroix, and Sayed}]{Jiang2024MixtralOE}
Albert~Q. Jiang, Alexandre Sablayrolles, Antoine Roux, Arthur Mensch, Blanche Savary, Chris Bamford, Devendra~Singh Chaplot, Diego de~Las~Casas, Emma~Bou Hanna, Florian Bressand, Gianna Lengyel, Guillaume Bour, Guillaume Lample, L'elio~Renard Lavaud, Lucile Saulnier, Marie-Anne Lachaux, Pierre Stock, Sandeep Subramanian, Sophia Yang, Szymon Antoniak, Teven~Le Scao, Th{\'e}ophile Gervet, Thibaut Lavril, Thomas Wang, Timoth{\'e}e Lacroix, and William~El Sayed. 2024.
\newblock \href {https://api.semanticscholar.org/CorpusID:266844877} {Mixtral of experts}.
\newblock \emph{ArXiv}, abs/2401.04088.

\bibitem[{Jiang et~al.(2023)Jiang, Sablayrolles, Mensch, Bamford, Chaplot, de~Las~Casas, Bressand, Lengyel, Lample, Saulnier, Lavaud, Lachaux, Stock, Scao, Lavril, Wang, Lacroix, and Sayed}]{Jiang2023Mistral7}
Albert~Qiaochu Jiang, Alexandre Sablayrolles, Arthur Mensch, Chris Bamford, Devendra~Singh Chaplot, Diego de~Las~Casas, Florian Bressand, Gianna Lengyel, Guillaume Lample, Lucile Saulnier, L'elio~Renard Lavaud, Marie-Anne Lachaux, Pierre Stock, Teven~Le Scao, Thibaut Lavril, Thomas Wang, Timoth{\'e}e Lacroix, and William~El Sayed. 2023.
\newblock \href {https://api.semanticscholar.org/CorpusID:263830494} {Mistral 7b}.
\newblock \emph{ArXiv}, abs/2310.06825.

\bibitem[{Lev et~al.(2019)Lev, Shmueli-Scheuer, Herzig, Jerbi, and Konopnicki}]{Lev2019TalkSummAD}
Guy Lev, Michal Shmueli-Scheuer, Jonathan Herzig, Achiya Jerbi, and David Konopnicki. 2019.
\newblock \href {https://doi.org/10.18653/v1/P19-1204} {{T}alk{S}umm: A dataset and scalable annotation method for scientific paper summarization based on conference talks}.
\newblock In \emph{Proceedings of the 57th Annual Meeting of the Association for Computational Linguistics}, pages 2125--2131, Florence, Italy. Association for Computational Linguistics.

\bibitem[{Li et~al.(2021)Li, Huang, Ma, and Lin}]{Li2021TowardsTS}
Da-Wei Li, Danqing Huang, Tingting Ma, and Chin-Yew Lin. 2021.
\newblock \href {https://api.semanticscholar.org/CorpusID:235363459} {Towards topic-aware slide generation for academic papers with unsupervised mutual learning}.
\newblock In \emph{AAAI Conference on Artificial Intelligence}.

\bibitem[{Lin(2004)}]{Lin2004ROUGEAP}
Chin-Yew Lin. 2004.
\newblock \href {https://api.semanticscholar.org/CorpusID:964287} {Rouge: A package for automatic evaluation of summaries}.
\newblock In \emph{Annual Meeting of the Association for Computational Linguistics}.

\bibitem[{Lukas et~al.(2023)Lukas, Salem, Sim, Tople, Wutschitz, and Zanella-B'eguelin}]{Lukas2023AnalyzingLO}
Nils Lukas, A.~Salem, Robert Sim, Shruti Tople, Lukas Wutschitz, and Santiago Zanella-B'eguelin. 2023.
\newblock \href {https://api.semanticscholar.org/CorpusID:256459554} {Analyzing leakage of personally identifiable information in language models}.
\newblock \emph{2023 IEEE Symposium on Security and Privacy (SP)}, pages 346--363.

\bibitem[{Manakul et~al.(2023)Manakul, Liusie, and Gales}]{Manakul2023SelfCheckGPTZB}
Potsawee Manakul, Adian Liusie, and Mark John~Francis Gales. 2023.
\newblock \href {https://api.semanticscholar.org/CorpusID:257557820} {Selfcheckgpt: Zero-resource black-box hallucination detection for generative large language models}.
\newblock \emph{ArXiv}, abs/2303.08896.

\bibitem[{Papineni et~al.(2002)Papineni, Roukos, Ward, and Zhu}]{Papineni2002BleuAM}
Kishore Papineni, Salim Roukos, Todd Ward, and Wei-Jing Zhu. 2002.
\newblock \href {https://api.semanticscholar.org/CorpusID:11080756} {Bleu: a method for automatic evaluation of machine translation}.
\newblock In \emph{Annual Meeting of the Association for Computational Linguistics}.

\bibitem[{Qiang et~al.(2016)Qiang, Fu, Guo, Zhou, and Sigal}]{Qiang2016LearningTG}
Yuting Qiang, Yanwei Fu, Yanwen Guo, Zhi-Hua Zhou, and Leonid Sigal. 2016.
\newblock \href {https://api.semanticscholar.org/CorpusID:9123423} {Learning to generate posters of scientific papers}.
\newblock \emph{ArXiv}, abs/1604.01219.

\bibitem[{Radford et~al.(2019)Radford, Wu, Child, Luan, Amodei, and Sutskever}]{Radford2019LanguageMA}
Alec Radford, Jeff Wu, Rewon Child, David Luan, Dario Amodei, and Ilya Sutskever. 2019.
\newblock \href {https://api.semanticscholar.org/CorpusID:160025533} {Language models are unsupervised multitask learners}.

\bibitem[{Scir{\'e} et~al.(2023)Scir{\'e}, Conia, Ciciliano, and Navigli}]{Scir2023EchoesFA}
Alessandro Scir{\'e}, Simone Conia, Simone Ciciliano, and Roberto Navigli. 2023.
\newblock \href {https://api.semanticscholar.org/CorpusID:259095692} {Echoes from alexandria: A large resource for multilingual book summarization}.
\newblock In \emph{Annual Meeting of the Association for Computational Linguistics}.

\bibitem[{See et~al.(2017)See, Liu, and Manning}]{see-etal-2017-get}
Abigail See, Peter~J. Liu, and Christopher~D. Manning. 2017.
\newblock \href {https://doi.org/10.18653/v1/P17-1099} {Get to the point: Summarization with pointer-generator networks}.
\newblock In \emph{Proceedings of the 55th Annual Meeting of the Association for Computational Linguistics (Volume 1: Long Papers)}, pages 1073--1083, Vancouver, Canada. Association for Computational Linguistics.

\bibitem[{Sefid and Giles(2022)}]{Sefid2022SciBERTSUMES}
Athar Sefid and C.~Lee Giles. 2022.
\newblock \href {https://doi.org/10.1007/978-3-031-06555-2_46} {Scibertsum: Extractive summarization for scientific documents}.
\newblock In \emph{Document Analysis Systems: 15th IAPR International Workshop, DAS 2022, La Rochelle, France, May 22–25, 2022, Proceedings}, page 688–701, Berlin, Heidelberg. Springer-Verlag.

\bibitem[{Sun et~al.(2021)Sun, Hou, Wang, Zhang, and Wang}]{Sun2021D2SDG}
Edward Sun, Yufang Hou, Dakuo Wang, Yunfeng Zhang, and Nancy X.~R. Wang. 2021.
\newblock \href {https://doi.org/10.18653/v1/2021.naacl-main.111} {{D}2{S}: Document-to-slide generation via query-based text summarization}.
\newblock In \emph{Proceedings of the 2021 Conference of the North American Chapter of the Association for Computational Linguistics: Human Language Technologies}, pages 1405--1418, Online. Association for Computational Linguistics.

\bibitem[{Szmidt and Kacprzyk(2010)}]{5548399}
Eulalia Szmidt and Janusz Kacprzyk. 2010.
\newblock \href {https://doi.org/10.1109/IS.2010.5548399} {The spearman rank correlation coefficient between intuitionistic fuzzy sets}.
\newblock In \emph{2010 5th IEEE International Conference Intelligent Systems}, pages 276--280.

\bibitem[{Vasilyev et~al.(2020)Vasilyev, Dharnidharka, and Bohannon}]{Vasilyev2020FillIT}
Oleg Vasilyev, Vedant Dharnidharka, and John Bohannon. 2020.
\newblock \href {https://doi.org/10.18653/v1/2020.eval4nlp-1.2} {Fill in the {BLANC}: Human-free quality estimation of document summaries}.
\newblock In \emph{Proceedings of the First Workshop on Evaluation and Comparison of NLP Systems}, pages 11--20, Online. Association for Computational Linguistics.

\bibitem[{Vaswani et~al.(2017)Vaswani, Shazeer, Parmar, Uszkoreit, Jones, Gomez, Kaiser, and Polosukhin}]{Vaswani2017AttentionIA}
Ashish Vaswani, Noam Shazeer, Niki Parmar, Jakob Uszkoreit, Llion Jones, Aidan~N Gomez, \L~ukasz Kaiser, and Illia Polosukhin. 2017.
\newblock \href {https://proceedings.neurips.cc/paper_files/paper/2017/file/3f5ee243547dee91fbd053c1c4a845aa-Paper.pdf} {Attention is all you need}.
\newblock In \emph{Advances in Neural Information Processing Systems}, volume~30. Curran Associates, Inc.

\bibitem[{Wang et~al.(2023{\natexlab{a}})Wang, Liu, Liu, Neshati, Ma, Zhu, and Zhao}]{Wang2023Slide4NCP}
Fengjie Wang, Xuye Liu, Oujing Liu, Ali Neshati, Tengfei Ma, Min Zhu, and J.~Zhao. 2023{\natexlab{a}}.
\newblock \href {https://api.semanticscholar.org/CorpusID:258216753} {Slide4n: Creating presentation slides from computational notebooks with human-ai collaboration}.
\newblock \emph{Proceedings of the 2023 CHI Conference on Human Factors in Computing Systems}.

\bibitem[{Wang et~al.(2023{\natexlab{b}})Wang, Wei, Schuurmans, Le, Chi, Narang, Chowdhery, and Zhou}]{Wang2022SelfConsistencyIC}
Xuezhi Wang, Jason Wei, Dale Schuurmans, Quoc~V Le, Ed~H. Chi, Sharan Narang, Aakanksha Chowdhery, and Denny Zhou. 2023{\natexlab{b}}.
\newblock \href {https://openreview.net/forum?id=1PL1NIMMrw} {Self-consistency improves chain of thought reasoning in language models}.
\newblock In \emph{The Eleventh International Conference on Learning Representations}.

\bibitem[{Wang et~al.(2015)Wang, Kawai, and Sumiya}]{Wang2015iPosterIP}
Yuanyuan Wang, Yukiko Kawai, and Kazutoshi Sumiya. 2015.
\newblock \href {https://api.semanticscholar.org/CorpusID:55145092} {iposter: Interactive poster generation based on topic structure and slide presentation}.
\newblock \emph{Transactions of The Japanese Society for Artificial Intelligence}, 30:112--123.

\bibitem[{Wei et~al.(2022{\natexlab{a}})Wei, Tay, Bommasani, Raffel, Zoph, Borgeaud, Yogatama, Bosma, Zhou, Metzler, Chi, Hashimoto, Vinyals, Liang, Dean, and Fedus}]{Wei2022EmergentAO}
Jason Wei, Yi~Tay, Rishi Bommasani, Colin Raffel, Barret Zoph, Sebastian Borgeaud, Dani Yogatama, Maarten Bosma, Denny Zhou, Donald Metzler, Ed~H. Chi, Tatsunori Hashimoto, Oriol Vinyals, Percy Liang, Jeff Dean, and William Fedus. 2022{\natexlab{a}}.
\newblock \href {https://openreview.net/forum?id=yzkSU5zdwD} {Emergent abilities of large language models}.
\newblock \emph{Transactions on Machine Learning Research}.
\newblock Survey Certification.

\bibitem[{Wei et~al.(2022{\natexlab{b}})Wei, Wang, Schuurmans, Bosma, ichter, Xia, Chi, Le, and Zhou}]{Wei2022ChainOT}
Jason Wei, Xuezhi Wang, Dale Schuurmans, Maarten Bosma, brian ichter, Fei Xia, Ed~Chi, Quoc~V Le, and Denny Zhou. 2022{\natexlab{b}}.
\newblock \href {https://proceedings.neurips.cc/paper_files/paper/2022/file/9d5609613524ecf4f15af0f7b31abca4-Paper-Conference.pdf} {Chain-of-thought prompting elicits reasoning in large language models}.
\newblock In \emph{Advances in Neural Information Processing Systems}, volume~35, pages 24824--24837. Curran Associates, Inc.

\bibitem[{Xu and Wan(2021)}]{Xu2021NeuralCE}
Sheng Xu and Xiaojun Wan. 2021.
\newblock \href {https://api.semanticscholar.org/CorpusID:245218732} {Neural content extraction for poster generation of scientific papers}.
\newblock \emph{ArXiv}, abs/2112.08550.

\bibitem[{Yuan et~al.(2021)Yuan, Neubig, and Liu}]{Yuan2021BARTScoreEG}
Weizhe Yuan, Graham Neubig, and Pengfei Liu. 2021.
\newblock \href {https://openreview.net/forum?id=5Ya8PbvpZ9} {{BARTS}core: Evaluating generated text as text generation}.
\newblock In \emph{Advances in Neural Information Processing Systems}.

\bibitem[{Zdravkova et~al.(2023)Zdravkova, Dalipi, and Ahlgren}]{Zdravkova2023IntegrationOL}
Katerina Zdravkova, Fisnik Dalipi, and Fredrik Ahlgren. 2023.
\newblock \href {https://api.semanticscholar.org/CorpusID:267576205} {Integration of large language models into higher education: A perspective from learners}.
\newblock \emph{2023 International Symposium on Computers in Education (SIIE)}, pages 1--6.

\bibitem[{Zhang* et~al.(2020)Zhang*, Kishore*, Wu*, Weinberger, and Artzi}]{Zhang2019BERTScoreET}
Tianyi Zhang*, Varsha Kishore*, Felix Wu*, Kilian~Q. Weinberger, and Yoav Artzi. 2020.
\newblock \href {https://openreview.net/forum?id=SkeHuCVFDr} {Bertscore: Evaluating text generation with bert}.
\newblock In \emph{International Conference on Learning Representations}.

\bibitem[{Zhao et~al.(2019)Zhao, Peyrard, Liu, Gao, Meyer, and Eger}]{Zhao2019MoverScoreTG}
Wei Zhao, Maxime Peyrard, Fei Liu, Yang Gao, Christian~M. Meyer, and Steffen Eger. 2019.
\newblock \href {https://api.semanticscholar.org/CorpusID:202540033} {Moverscore: Text generation evaluating with contextualized embeddings and earth mover distance}.
\newblock In \emph{Conference on Empirical Methods in Natural Language Processing}.

\bibitem[{Zhong et~al.(2022)Zhong, Liu, Yin, Mao, Jiao, Liu, Zhu, Ji, and Han}]{Zhong2022TowardsAU}
Ming Zhong, Yang Liu, Da~Yin, Yuning Mao, Yizhu Jiao, Peng Liu, Chenguang Zhu, Heng Ji, and Jiawei Han. 2022.
\newblock \href {https://api.semanticscholar.org/CorpusID:252873117} {Towards a unified multi-dimensional evaluator for text generation}.
\newblock In \emph{Conference on Empirical Methods in Natural Language Processing}.

\end{thebibliography}
\bibliographystyle{acl_natbib}
\clearpage
\newpage
\appendix
\begin{table*}
\small
\centering
    \begin{tabular}{l|l}
    Prompt Function & Prompt \\ \hline
    Generate the intermediate representation & \begin{tabular}[c]{@{}l@{}}"Given the input text, extract the document title and authors.\\ For each section in the given input text, extract the most important sentences.\\ Format the output using the following JSON template:\textbackslash{}n\\ \textless{}SURe STRUCTURE\textgreater{}\textbackslash{}n\textbackslash{}n\\ Input: \textless{}INPUT DOCUMENT\textgreater \textbackslash{}n\\ Output:"\end{tabular} \\ \hdashline[0.5pt/3pt]
    Generate LaTeX document & \begin{tabular}[c]{@{}l@{}}"Summarize the following input in a \textless{}TEMPLATE TYPE\textgreater style.\\ Style parameters: \textless{}STYLE PARAMETERS\textgreater\\ Format the output document as a latex file:\textbackslash{}n\\ Input: \textless{}INPUT DOCUMENT\textgreater{}\textbackslash{}n\textbackslash{}n\\ Output:"\end{tabular}
    \end{tabular}
    \caption{Prompts used to generate the intermediate representation and final LaTeX document. The JSON structure is pictured in Figure~\ref{fig:sure-template}.} %\isabel{move to appendix if space is needed}}
    \label{tbl:prompts}
\end{table*}

\section{Prompt details}\label{sec:prompt-details}
\begin{figure}[t]
    \centering
\includegraphics[width=.9\columnwidth]{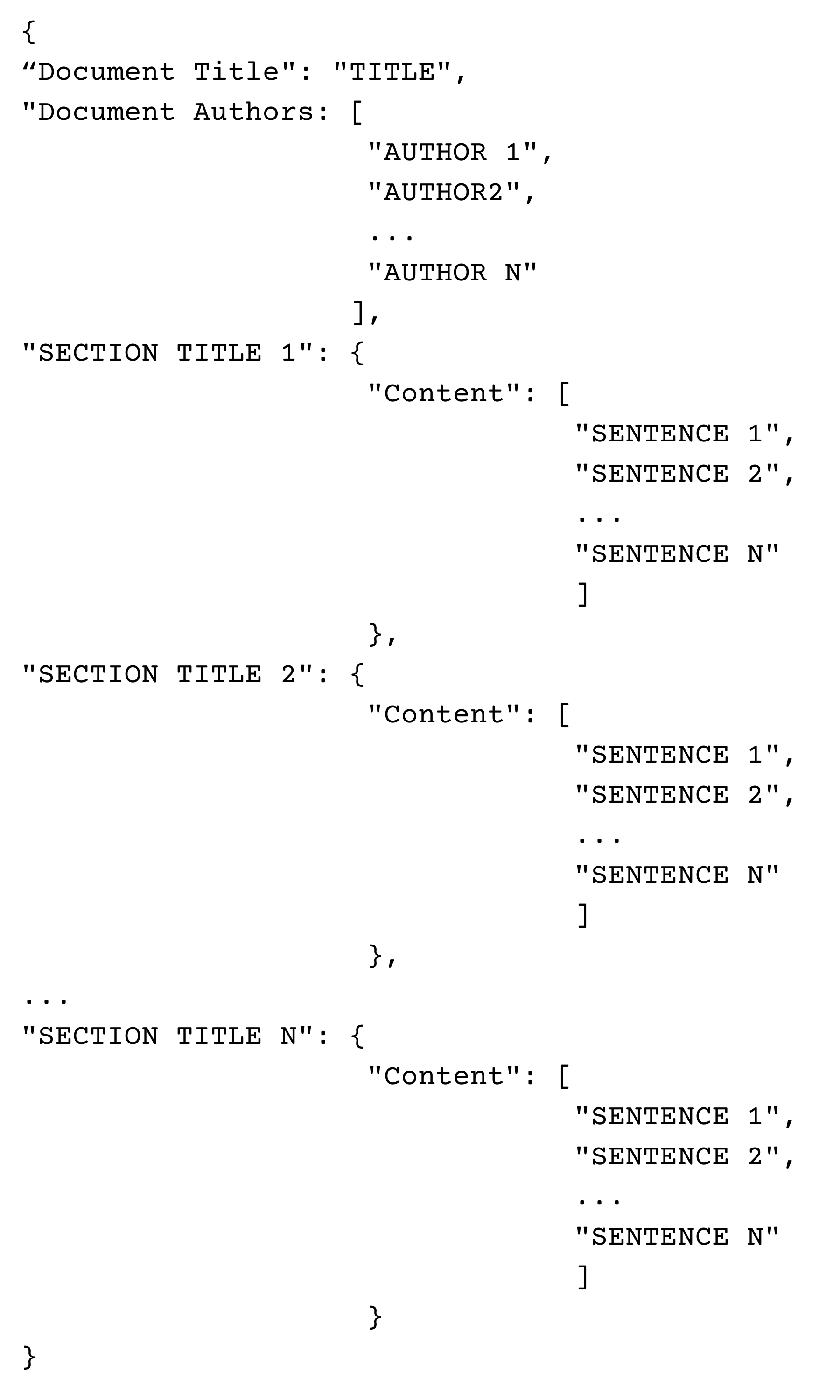}
    \caption{Template of the intermediate representation provided to the prompts in Table~\ref{tbl:prompts}.}
    \label{fig:sure-template}
\end{figure}

The prompts and intermediate representation template used can be found in Table~\ref{tbl:prompts} and Figure~\ref{fig:sure-template}, respectively. We note that the specific structure provided to the prompt is not inherent to our method, and a different structure could be provided depending on the input document and domain.
For the tasks we evaluate in this paper, we use the following style parameters:
\begin{itemize}
    \item \textbf{Slides:} ``Slides should include a title page. Following slides should contain an informative slide title and short, concise bullet points. Longer slides should be broken up into multiple slides.''
    \item \textbf{Posters:} ``Posters should include a title section at the top. Each panel should include a heading and short, concise bullet points of the most important take-aways from that section.''
    \item \textbf{Blogs:} ``Blogs should include paragraphs introducing the topic, a summary of the input document, and important takeaways. The blog should be more readable to a general audience than the input document.''
\end{itemize}

\section{Annotation Instructions}
\label{sec:annotation-instructions}
\subsection{Questions}
\textbf{Question 1 – Which document do you prefer?}
In this question, you are asked to choose which document version you prefer. Some examples of qualities you may use to decide your preference include: 
\begin{itemize}
    \item The quality of the content – The text is grammatical and understandable. E.g. Document A contains major grammatical errors while Document B only contains minor errors.
    \item The formatting – The formatting is reasonable and matches the formatting of the intended document type. E.g. A poster contains panels and each panel contains a header and body text.
    \item The style – The document matches the style of the intended document type. E.g. Shorter, bulleted sentences in a slide deck.
    \item Information represented in the document – The document contains sufficient information to represent the input document. E.g. A blog post represents the most important sections from the input document.
\end{itemize}

The above criteria are non-exhaustive. Not all criteria must be met, and you may use other relevant criteria to make your decision. 
You are not rating the document for factual correctness,\footnote{The annotators are non-experts and do not have the background to determine factual correctness of scientific information. Instead, they are encouraged to use the original paper to understand if the information presented in the documents represent the information in the paper, to the best of their understanding.} and only need to refer to the corresponding scientific article if it will aid in making your preference. 
You can answer this question with either Document A or Document B.

\textbf{Question 2 – On a scale of 1-3, to what degree do you prefer your selection?
}

\noindent In this question you will rate the degree to which you prefer your selection, on the following scale:
\begin{enumerate}
    \item Small preference – The documents are similar in quality and only contain minor differences that affect my preference.
    \item Moderate preference – I clearly prefer one document but the differences are not major.
    \item Strong preference – I have a strong preference for one document and the differences between the documents are major.
\end{enumerate}

\textbf{Question 3 – Why do you prefer your selection? (You may select more than one property)}
\begin{itemize}
    \item[$\square$] Formatting
    \item[$\square$] Information
    \item[$\square$] Quality
    \item[$\square$] Style
    \item[$\square$] Other (free text)
\end{itemize}

\subsection{Edge cases}
For most edge cases, it is up to your discretion on how to best handle the case. However, below are a few examples of how you could consider certain edge cases:

\textbf{Example 1: Slides 1-5 of Document A are higher quality but slides 6-10 of Document B are higher quality.
}
You could reason that the first slides represent the most important information, and choose Document A. However, since Document B contained higher quality slides for another portion of the document, you could rate your degree of preference as “Small preference.”

\textbf{Example 2: Document A more closely matches the style of the intended document type, but Document B contains more relevant information to the source document.
}
You could consider if Document A contains sufficient information to represent the input document, such as representing the most important sections. If yes, then you could prefer Document A. If not, then you could reason that information content is more important than style, and prefer Document B.

\textbf{Example 3: Document A contains more relevant information than Document B, but also contains major formatting errors, such as text being cut off from the document.}

You could reason that although Document A contains more relevant information, the major formatting errors are significant enough to prefer Document B.

\textbf{Example 4: Neither document matches the style or formatting of the intended document type. 
}
Since neither document matches the style or formatting of the intended document type, you could consider other criteria, such as quality of content or the information represented.

% \usepackage{multirow}
\begin{table}[t!]
\centering
\small
\begin{tabular}{ll|llll}
 &  & \multicolumn{4}{c}{Simularity Measure} \\
 & Temp  & R-L & M & B & BERTS \\ \hline
\multirow{5}{*}{\rotatebox[origin=c]{90}{Slides}} & 0.0 & 7.3 & 8.2 & \textbf{1.6} & 35.1 \\
 & 0.25  & 7.2 &  \textbf{8.3} & 1.5 & 35.4 \\
 & 0.5 & 7.0 & \textbf{8.3} & 1.5 & \textbf{36.4} \\
 & 0.75  & 7.0 &  8.0 & 1.2 &35.5 \\
 & 1.0  & \textbf{7.4} &  8.2 & 1.4 & 35.8 \\ \hdashline[0.5pt/3pt]
\multirow{5}{*}{\rotatebox[origin=c]{90}{Blogs}} & 0.0  & 25.3 &  19.9 & 2.7 & 82.7 \\
 & 0.25 & 25.5 &  19.8 & 2.7 & 82.6 \\
 & 0.5  & \textbf{26.2} & \textbf{20.9} & \textbf{3.2} & \textbf{82.8} \\
 & 0.75 & 25.4 & 19.9 & 2.7 & 82.6 \\
 & 1.0 &  24.8 & 19.9 & 2.6 & 82.7 \\ \hdashline[0.5pt/3pt]
\multirow{5}{*}{\rotatebox[origin=c]{90}{Posters}} & 0.0 & \textbf{13.5} &  \textbf{15.3} & \textbf{3.4} & \textbf{53.5} \\
 & 0.25 & 13.0 &  14.8 & \textbf{3.4} & 53.1 \\
 & 0.5 &  12.5 &  14.0 & 2.7 & 52.2 \\
 & 0.75 &  12.0 & 13.9 & 3.0 & 50.8 \\
 & 1.0 &  11.6 &  11.9 & 2.4 & 50.3
\end{tabular}
\caption{Results of the temperature hyperparameter experiments. We use ROUGE-L (R-L), METEOR (M), BLEU (B) and BERTScore (BERTS) as our similarity measures.}
\label{tbl:temp-exp}
\end{table}
\section{Temperature Experiments}\label{sec:temp-exp}
We experiment with the temperature of the generations to see how temperature affects performance. We randomly sample 100 documents each from LongSumm and Doc2PPT for the blog and poster generation tasks, respectively. We use the entirety of the Paper-Poster dataset, since it contains less than 100 examples. We use \texttt{gpt35-16k} and experiment with the temperatures $[0.0, 0.25, 0.5, 0.75, 1.0]$. The results of this experiment can be found in Table~\ref{tbl:temp-exp}. As we can see from the results, there seems to be little consistency across the different types in which temperature performs the best.

\section{Longsumm Blind Test Set Results}\label{sec:longsumm-blind}
We submit the final documents from GPT4, the best performing model overall, to the Longsumm blind test set evaluation. We compare the documents generated with and without the intermediate step. We see that without the intermediate representation we get a Rouge-1 score of 46.8 while the results generated without the intermediate representation received a Rouge-1 score of 46.4. We note that this blind test set of 22 papers is significantly smaller than the evaluation data (505 papers) we used in the main body of this paper. Despite not designing a task specific method, we place second on the leaderboard, showing the powerful capabilities of LLMs in long document generation.

\section{Example Outputs}\label{sec:example-outputs}
\begin{figure*}[t!]
    \centering
    \begin{subfigure}[t]{0.5\textwidth}
        \centering
        \frame{\includegraphics[width=.95\textwidth]{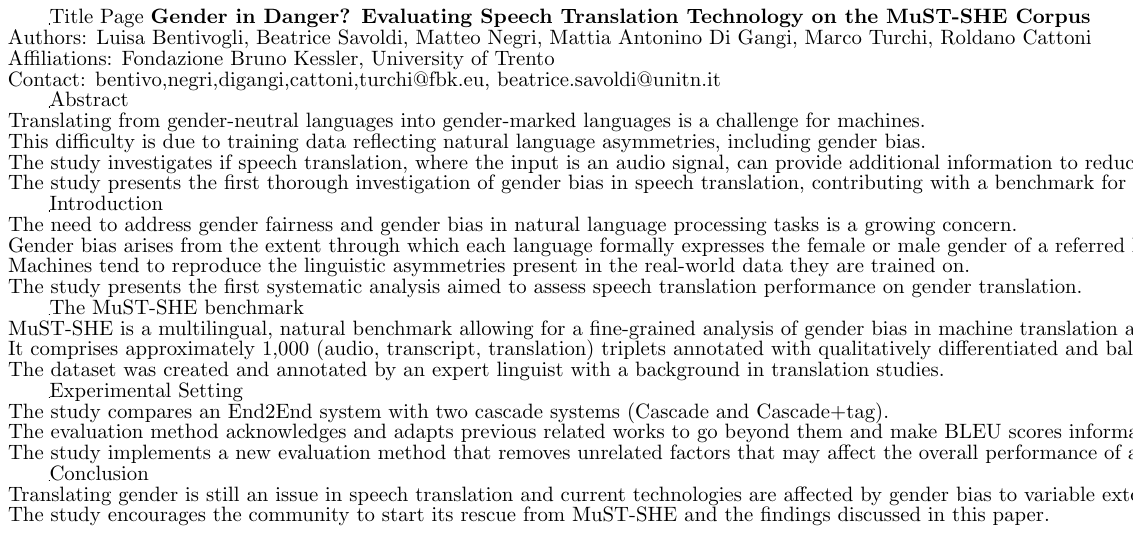}}
        \caption{Document generated without intermediate representation. This example is not cropped.}
    \end{subfigure}%
    ~ 
    \begin{subfigure}[t]{0.5\textwidth}
        \centering
        \frame{\includegraphics[width=.9\textwidth]{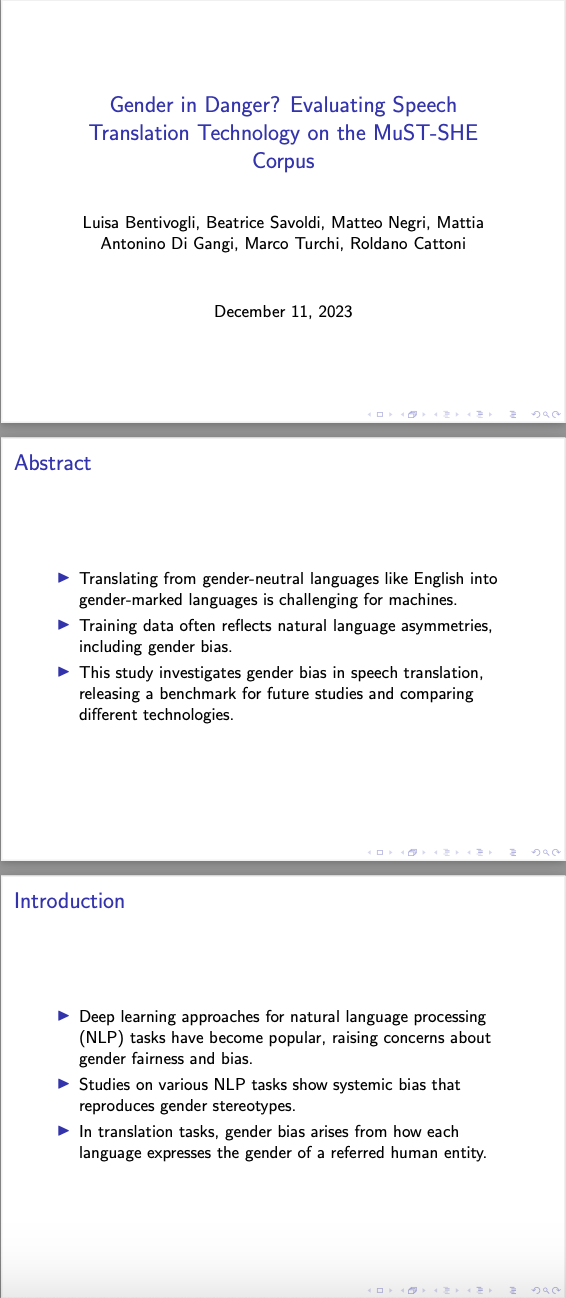}}
        \caption{Document generated with intermediate representation. This example is cropped for space and includes an additional 4 slides that are not included for space.}
    \end{subfigure}
    \caption{The above documents are example slides generated by GPT4 ($\texttt{gpt4-32k}$) with and without the intermediate representation. We can see that without the intermediate step, the model did not generate a true slide deck.}
    \label{fig:ex-slides}
\end{figure*}

\begin{figure*}[t!]
    \centering
    \begin{subfigure}[t]{0.5\textwidth}
        \centering
        \frame{\includegraphics[width=.95\textwidth]{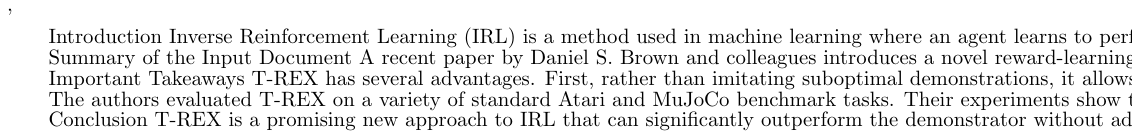}}
        \caption{Document generated without the intermediate representation. This example is not cropped.}
    \end{subfigure}%
    ~ 
    \begin{subfigure}[t]{0.5\textwidth}
        \centering
        \frame{\includegraphics[width=.95\textwidth]{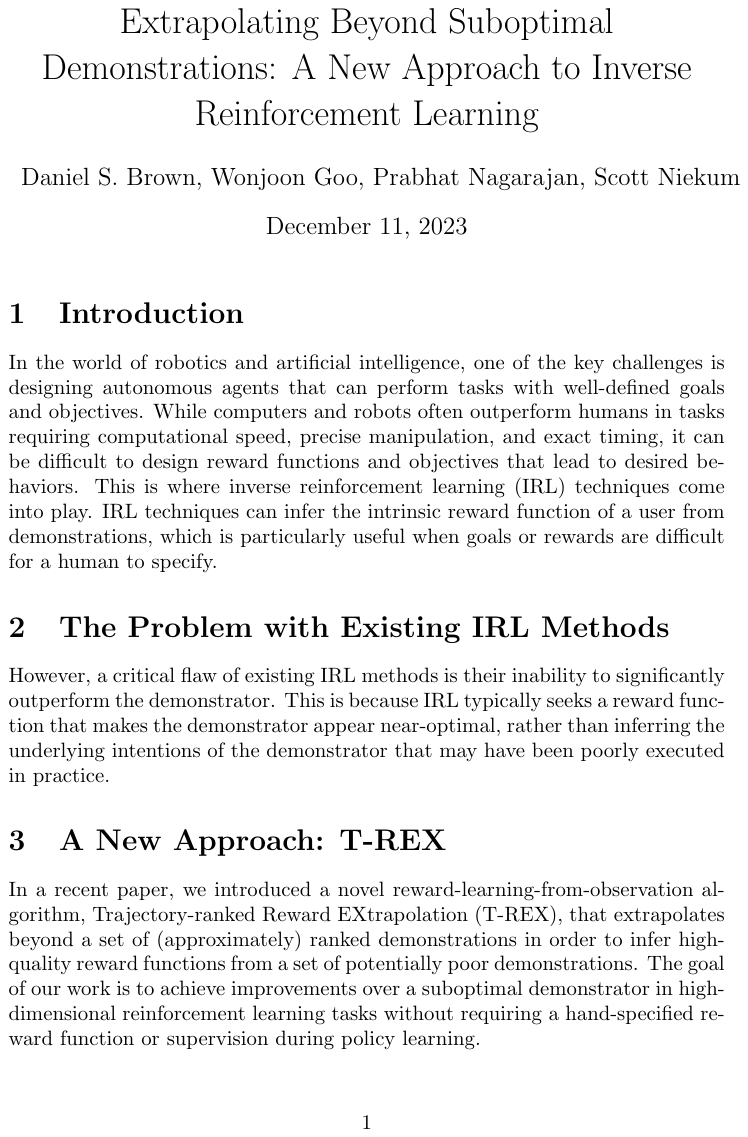}}
        \caption{Document generated with the intermediate representation. This example is cropped for space and includes an additional page of text.}
    \end{subfigure}
    \caption{The above documents are example blog posts generated by GPT4 ($\texttt{gpt4-32k}$) with and without the intermediate representation. We can see that without the intermediate representation, the model did not properly format the LaTeX file for compilation.}
    \label{fig:ex-blogs}
\end{figure*}

\begin{figure*}[t!]
    \centering
    \begin{subfigure}[t]{0.5\textwidth}
        \centering
        \frame{\includegraphics[width=.95\textwidth]{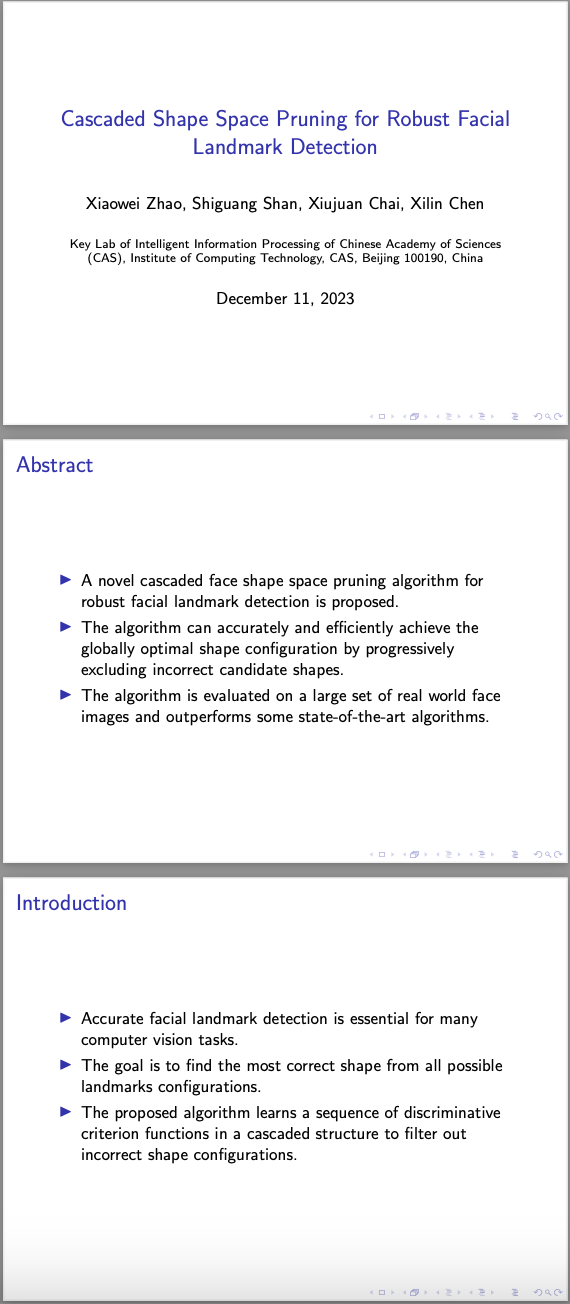}}
        \caption{Document generated without the intermediate representation. This example is cropped for space and includes an additional 3 slides.}
    \end{subfigure}%
    ~ 
    \begin{subfigure}[t]{0.5\textwidth}
        \centering
        \frame{\includegraphics[width=.95\textwidth]{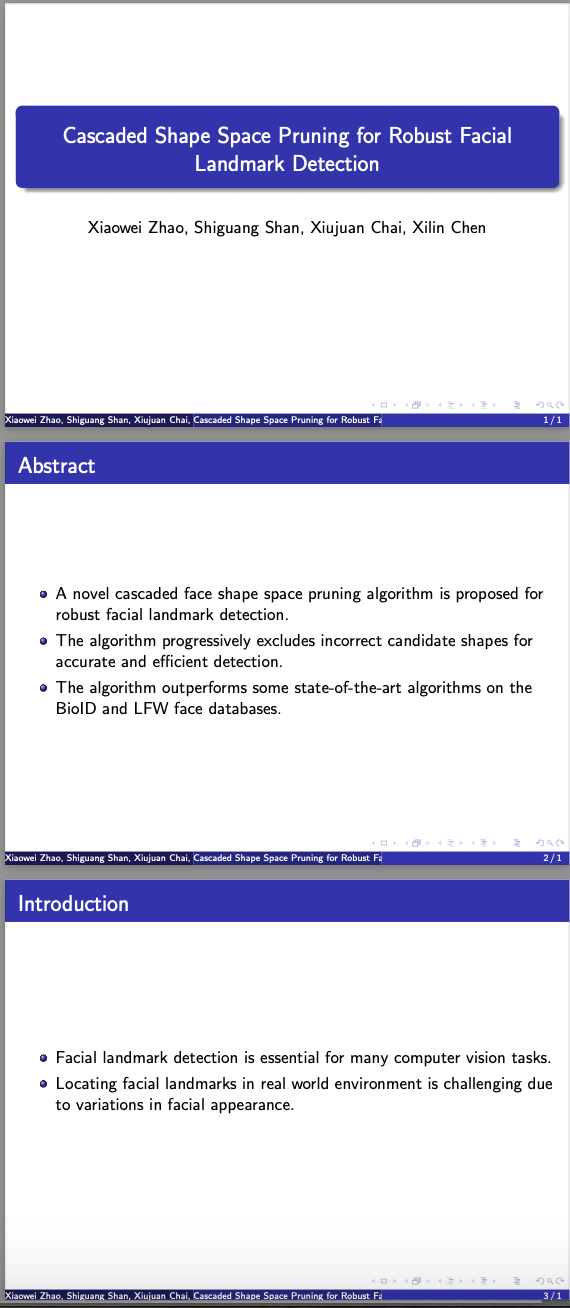}}
        \caption{Document generated with the intermediate representation. This example is cropped for space and includes an additional 4 slides.}
    \end{subfigure}
    \caption{The above documents are example posters generated by GPT4 ($\texttt{gpt4-32k}$) with and without the intermediate representation. We found that GPT4 often generates slide decks in place of posters. We can see that the document generated without the intermediate representation contains more verbose panels and includes less formatting.}
    \label{fig:ex-posters}
\end{figure*}

We provide examples of the outputs generated with and without the intermediate representation below. The documents in all examples are generated with GPT4 ($\texttt{gpt4-32k}$). Figure~\ref{fig:ex-slides} includes example slide generations, Figure~\ref{fig:ex-blogs} includes example blog generations, and Figure~\ref{fig:ex-posters} includes example poster generations. 

\end{document}